\documentclass[doublecolumn]{IEEEtran}
\usepackage{ragged2e}
\usepackage{cite}
\usepackage{cuted}
\usepackage{caption}
\usepackage{comment}
\usepackage{multirow}
\usepackage{makecell}
\ifCLASSOPTIONcompsoc
\usepackage[caption=false, font=normalsize, labelfont=sf, textfont=sf]{subfig}
\else
\usepackage[caption=false, font=footnotesize]{subfig}

\ifCLASSINFOpdf
  \usepackage[pdftex]{graphicx}
\else
  \usepackage[dvips]{graphicx}
\fi
\usepackage{amsmath}
\usepackage{amssymb}
\usepackage{amsthm}
\makeatletter

\newcommand{\Rmnum}[1]{\expandafter\@slowromancap\romannumeral #1@}
\makeatother

\newtheorem{theorem}{Theorem}
\newtheorem{lemma}{Lemma}

\newtheorem{assumption}{Assumption}
\newtheorem{definition}{Definition}
\newtheorem{proposition}{Proposition}

\usepackage{graphicx}

\usepackage{extarrows}
\usepackage{float}
\usepackage{bm}
\usepackage{algorithmic}
\usepackage{array}
\usepackage[table]{xcolor}
\usepackage{booktabs}
\usepackage{algorithm}
\usepackage{algorithmic}
\usepackage{ wasysym }

\usepackage{threeparttable}

\newenvironment{shrinkfix}
{ \bgroup
\addtolength\abovedisplayshortskip{-0.2ex}
\addtolength\abovedisplayskip{-0.2ex}
\addtolength\belowdisplayshortskip{-0.2ex}
\addtolength\belowdisplayskip{-0.2ex}}
{\egroup\ignorespacesafterend}

\ifCLASSOPTIONcompsoc
  \usepackage[caption=false,font=normalsize,labelfont=sf,textfont=sf]{subfig}
\else
  \usepackage[caption=false,font=footnotesize]{subfig}
\fi

\ifCLASSOPTIONcaptionsoff
 \usepackage[nomarkers]{endfloat}
 \let\MYoriglatexcaption\caption
 \renewcommand{\caption}[2][\relax]{\MYoriglatexcaption[#2]{#2}}
\fi

\usepackage{url}

\allowdisplaybreaks[4]

\begin{document}

%\title{Learn time-varying graphs with network temporal structures}
    \title{Graph Learning Across Data Silos}
\author{
Xiang~Zhang,~\IEEEmembership{Student~Member,~IEEE,} ~Qiao~Wang,~\IEEEmembership{Senior~Member,~IEEE}

\thanks{The authors are with the School of Information Science and Engineering, Southeast University, Nanjing 210096, China (e-mail: xiangzhang369@seu.edu.cn, 
qiaowang@seu.edu.cn).
 
}% <-this % stops a space

}

\maketitle

\begin{abstract}
We consider the problem of inferring graph topology from smooth graph signals in a novel but practical scenario where data are located in multiple clients and prohibited from leaving local clients due to factors such as privacy concerns. The main difficulty in this task is how to exploit the potentially heterogeneous data of all clients under data silos. To this end, we first propose an auto-weighted multiple graph learning model to jointly learn a personalized graph for each local client and a single consensus graph for all clients. The personalized graphs match local data distributions, thereby mitigating data heterogeneity, while the consensus graph captures the global information. Moreover, the model can automatically assign appropriate contribution weights to local graphs based on their similarity to the consensus graph. We next devise a  tailored algorithm to solve the induced problem, where all raw data are processed locally without leaving clients. Theoretically, we establish a provable estimation error bound and convergence analysis for the proposed model and algorithm. Finally, extensive experiments on synthetic and real data are carried out, and the results illustrate that our approach can learn graphs effectively in the target scenario.

\end{abstract}


\section{Introduction}
\label{sec:introduction}
%\subsection{Background and Motivation}
Graphs are powerful tools for flexibly describing topological relationships between data entities \cite{dong2019learning}, which have been extensively applied in many celebrated models, e.g., spectral clustering \cite{von2007tutorial} and graph neural networks (GNNs) \cite{wu2020comprehensive}. Among these applications, a graph that accurately represents the information inherent in the structured data is required, which, however, is not available in many cases. An alternative approach is to learn graphs directly from raw data, termed graph learning (GL), for downstream tasks \cite{mateos2019connecting, dong2019learning}.

In the literature, many studies learn graphs from statistical models, such as Gaussian Graphical Models (GGMs)\cite{friedman2008sparse}.  In general, these models aim to infer precision matrices (inverse covariance matrices), which encode conditional independence between random variables. Furthermore, \cite{egilmez2017graph, ying2020nonconvex, cai2024fast} introduce (generalized) Laplacian constraints on the learned precision matrices, which enjoy many downstream applications such as graph spectral analysis \cite{von2007tutorial}. Recently, with the rise of graph signal processing (GSP) \cite{ortega2018graph}, various methods attempted to learn graphs from a signal processing perspective. One of the most studied GSP-based models\textemdash the smoothness-based model\textemdash postulates that the observed graph signals are smooth over the underlying graph\cite{dong2016learning, kalofolias2016learn}. Intuitively, a smooth graph signal means that the signal values corresponding to two connected nodes of the underlying graph are similar\cite{kalofolias2016learn}. Typically, learning graphs from smooth signals is equivalent to minimizing a constrained quadratic Laplacian form problem, where the quadratic Laplacian form is related to signal smoothness, and the constraints (regularizers) are used to assign properties to the graphs, such as sparsity \cite{dong2016learning} and node connectivity \cite{kalofolias2016learn}. Many signals appear to be smooth over their underlying graphs, e.g.,  meteorology data \cite{dong2016learning} and medical data \cite{pu2021learning}, implying numerous applications of smoothness-based GL.

Here, we consider learning graphs in a previously unexplored scenario, where data are stored across multiple clients (e.g., companies and organizations) and are not allowed to leave their clients due to factors such as privacy concerns. This scenario is known as data-silos \cite{kairouz2021advances}, and a typical example is medical data. Suppose some hospitals collect brain fMRI data from autistic and non-autistic individuals separately to learn the impact of autism on connectivity networks of brain functional regions \cite{pu2021learning}. However, the data are prohibited from leaving the hospital where they are stored as people are reluctant to disclose their private data. A naive way is for each client to infer graph topology independently, which avoids any data leakage. Although feasible, this approach ignores potential topological relationships between local graphs, resulting in suboptimal results. Thus, it is more reasonable to learn graphs using data information from all clients. The task is not trivial and faces two main challenges. The first challenge is how to leverage siloed data from all clients collaboratively. Furthermore, graph signals across different silos are inherently non-IID due to factors such as different data collection protocols and heterogeneous underlying graphs \cite{rieke2020future}. We still use the brain fMRI data as an example. The brain functional connectivity networks of autistic and non-autistic individuals are different due to the impact of autism. It is unreasonable to utilize heterogeneous data to learn a single global graph like traditional paradigms \cite{dong2016learning,kalofolias2016learn}. Thus, the second challenge is handling heterogeneous data.

Regarding the first challenge, federated learning (FL)\cite{li2020federated, yang2019survey, kairouz2021advances} is an emerging tool for learning models based on datasets distributed across multiple clients under privacy constraints. The primary feature of FL is that clients transmit model updates instead of raw data to a central server to collaboratively learn a global model \cite{li2020federated}.  Privacy can be preserved in this schema to some extent since all data are processed locally \cite{yang2019survey}. However, traditional FL algorithms, e.g., FedAvg \cite{mcmahan2017communication}, learn a single global model from all data, which may suffer from performance degradation when data across different clients are heterogeneous. A widely used approach to handle data heterogeneity\textemdash the second challenge\textemdash is personalized FL  (PFL) \cite{tan2022towards}. The philosophy behind PFL is that we learn for each client a personalized model that matches its data distribution to mitigate the impact of heterogeneity.  The techniques for adapting global models for individual clients include transfer learning \cite{mansour2020three}, multi-task learning \cite{smith2017federated}, and meta-learning \cite{finn2017model}.

Borrowing the idea from PFL, we learn a personalized graph for each local client to handle data heterogeneity. However, it poses two additional challenges due to the characteristics of GL tasks. (\romannumeral1) Our goal is to learn all local graphs jointly by exploiting their latent relationships so that each local graph benefits from ``borrowing" information from other datasets. Thus, it is crucial to describe the relationships between local graphs, which is the focus of multiple graph learning (MGL). A common approach is to design regularization penalties, e.g., fused Lasso penalty \cite{danaher2014joint},  group Lasso penalty \cite{bickel2008regularized},  Gram matrix-based penalty \cite{yuan2021joint}, and those describing temporally topological relationships \cite{zhang2022time, yamada2019time, zhang2024}. These regularizers characterize topological relationships among multiple graphs, but few consider capturing common structures from all local graphs.  In practice, many graphs share common structures.  For example, connections between normally functioning brain regions in autistic and non-autistic individuals should remain the same. The common structures reflect the global information across all local datasets, which may be useful for many downstream tasks. (\romannumeral2) It is infeasible to apply existing PFL algorithms, such as \cite{t2020personalized, bellet2018personalized, smith2017federated, marfoq2021federated, chen2022personalized}, to our task since their personalization methods do not fit the GL problem. Besides, the problem of concern can be categorized as cross-silo FL, where clients are a few companies or organizations rather than massive devices in cross-device FL \cite{li2020federated, mcmahan2017communication}. Thus, a tailored algorithm is required to learn graphs in the target scenario.

To address these issues, we propose a framework to learn graphs from smooth but heterogeneous data under data silos. Our contributions can be summarized as follows:

\begin{enumerate}
\item[$\bullet$]
We propose an auto-weighted MGL model in which local personalized graphs and a consensus graph are jointly learned. The consensus graph can capture common structures representing global information across all datasets, while the local graphs preserve the heterogeneity of local datasets. Furthermore, our model can automatically assign contribution weights to local graphs based on their similarity to the consensus graph. Theoretically, we provide the estimation error bound of the proposed method to reveal some key factors affecting graph estimation performance.

\item[$\bullet$]
We develop a tailored algorithm to learn graphs under privacy constraints. Our algorithm follows the communication protocol of FL, where model updates instead of raw data are transmitted to a central server to learn all graphs collaboratively. The convergence analysis of the proposed algorithm is also provided.

\item[$\bullet$]
Extensive experiments with synthetic and real-world data are conducted to validate our framework, and the results show that our approach can effectively learn graphs under data silos.
\end{enumerate}

\textbf{Organization:} The rest of this paper is organized as follows.  We start with background information and the problem of concern in Section \ref{sec:background}. The proposed model for learning graphs in the target scenario is presented in \ref{sec:formulation-1}, followed by the corresponding algorithm in Section  \ref{sec:algorithm}. Experimental setups and results are provided in Section \ref{sec:Experiments}. Finally, concluding remarks are presented in Section \ref{sec:Conclusion}.

% \subsection{Notations}
\textbf{Notations:}
Throughout this paper, vectors, matrices, and sets are written in bold lowercase letters, bold uppercase letters, and calligraphic uppercase letters, respectively. Given a vector $\mathbf{y}$ and matrix $\mathbf{Y}$, $\mathbf{y}{[i]}$ and $\mathbf{Y}{[ij]}$ are the $i$-th entry of $\mathbf{y}$ and the $(i,j)$ entry of $\mathbf{Y}$. Besides,  $\mathbf{1}$, $\mathbf{0}$, and $\mathbf{I}$ represent all-one vectors, all-zero vectors, and identity matrices, respectively. For a vector or matrix, 
the $\ell_1$, $\ell_2$, $\ell_{\infty}$, and Frobenius norm are represented by $\lVert \cdot\rVert_1$, $\lVert \cdot \rVert_2$, $\lVert \cdot \rVert_{\infty}$, $\lVert \cdot \rVert_{\mathrm{F}}$,  respectively. Moreover, $\lVert \cdot \rVert_{\mathrm{F,off}}$ denotes the Frobenius norm of off-diagonal elements of a matrix, and $\mathrm{diag}(\cdot)$ means converting a vector to a diagonal matrix. The notations $\circ, \dag$, $\mathrm{Tr}(\cdot)$ stand for Hadamard product, pseudo inverse, and trace operator, respectively. For a set $\mathcal{Y}$, $\mathrm{conv}\left[\mathcal{Y}\right]$ is the affine and convex hulls of $\mathcal{Y}$. Finally, $\mathbb{R}$ and $\mathbb{S}$ represent the domain of real values and symmetric matrices whose dimensions depend on the context.

\begin{comment}

\section{Related Work}
\label{sec:related work}
\input{RelatedWork/relatedwork}
\end{comment}

\section{Background and Problem Statement}
\label{sec:background}

\subsection{GSP Background}
\label{sec:prem}
We consider undirected graphs with non-negative weights and no self-loops. For such a graph $\mathcal{G} = \{\mathcal{V},\mathcal{E}\}$ with $d$ vertices, where $\mathcal{V}$ and $\mathcal{E}$ are the sets of vertices and edges, respectively, its adjacency matrix  $\mathbf{A} \in \mathbb{S}^{d\times d}$ is a symmetric matrix with zero diagonal entries and non-negative off-diagonal entries. The Laplacian matrix  of $\mathcal{G}$ is $\mathbf{L} = \mathbf{D} - \mathbf{A}$  \cite{stankovic2019introduction}, where the degree matrix $\mathbf{D}\in\mathbb{S}^{d\times d}$ is a diagonal matrix satisfying $\mathbf{D}{[ii]} = \sum_{j=1}^d \mathbf{A}{[ij]}$. The matrices $\mathbf{A}$ and $\mathbf{L}$ encode the topology of $\mathcal{G}$ since they have a one-to-one relationship. We study the graph signal $\mathbf{x} = \left[\mathbf{x}{[1]},\dots,\mathbf{x}{[d]} \right]^{\top} \in \mathbb{R}^d$  associated with $\mathcal{G}$, where $\mathbf{x}{[i]}$ is the signal value of node $i\in\mathcal{V}$. The smoothness of $\mathbf{x}$ over $\mathcal{G}$ is defined as follows.

\begin{definition}
(Smoothness \cite{dong2016learning}). Given a graph signal $\mathbf{x}$ and a graph $\mathcal{G}$ whose Laplacian matrix and adjacency matrix are $\mathbf{L}$ and $\mathbf{A}$, respectively, the smoothness of $\mathbf{x}$ over $\mathcal{G}$ is 
\begin{shrinkfix}
    \begin{align}
    \mathbf{x }^{\top}\mathbf{L}\mathbf{x } = \frac{1}{2}\sum_{i,j}\mathbf{A}{[ij]}\left(\mathbf{x}{[i]}-\mathbf{x}{[j]}\right)^2.
    \label{smoothness}
    \end{align}
\end{shrinkfix}

\label{definition-smoothness}
\vspace{-0.5em}
\end{definition}
The Laplacian quadratic form  \eqref{smoothness} is known as the Dirichlet energy, which quantifies how much the signal $\mathbf{x}$ changes w.r.t. $\mathcal{G}$. A small value of \eqref{smoothness} indicates limited signal variability, meaning that $\mathbf{x }$ is smooth over the corresponding graph \cite{dong2016learning}. %From the definition, if a signal is smooth over $\mathcal{G}$, the values of two dimensions connected with large edges weight of $\mathcal{G}$ should be similar. 

\subsection{Graph Learning From Smooth Signals}
Given $N$ observations $\mathbf{X} = \left[\mathbf{x}_1,\dots,\mathbf{x}_N\right]\in\mathbb{R}^{d\times N}$, smoothness-based GL aims to infer the underlying graph topology $\mathcal{G}$ under the assumption that the signals $\mathbf{X}$ are smooth over $\mathcal{G}$.  Formally, the problem is written as 
\begin{shrinkfix}
    \begin{align}
    \underset{\mathbf{L}\in \mathcal{L}}{\mathrm{min}}\,\,\frac{1}{N}\sum_{n = 1}^{N} \mathbf{x}_n^{\top}\mathbf{L}\mathbf{x}_n - \alpha\mathbf{1}^{\top}\log\left(\mathrm{diag}(\mathbf{L})\right) + {\beta} \lVert \mathbf{L}\rVert^2_{\mathrm{F, off}},
    \label{GL-L-common}
    \end{align}
\end{shrinkfix}
where the first term is to quantify the smoothness of $\mathbf{X}$ over $\mathcal{G}$, and the last two terms are regularizers that endow $\mathcal{G}$ with desired properties.  The first and second terms of the regularizers control node connectivity and edge sparsity of the learned graph \cite{kalofolias2016learn}, where $\alpha$ and $\beta$ are predefined constants. We use Laplacian matrix $\mathbf{L}$ to represent graph topology of $\mathcal{G}$, which lies in the following set $\mathcal{L}$  
\begin{shrinkfix}
    \begin{align}
    \mathcal{L} \triangleq \left\{\mathbf{L}: \mathbf{L}\in \mathbb{S}^{d\times d},\, \mathbf{L}\mathbf{1} = \mathbf{0},\, \mathbf{L}{[ij]}\leq 0 \,\, \text{for}\,\, i \neq j \right\}.
    \label{LaplacianSets}
    \end{align}
\end{shrinkfix}
Based on \eqref{smoothness},    problem \eqref{GL-L-common} can be rephrased as
\begin{shrinkfix}
    \begin{align}
    \underset{\mathbf{A}\in \mathcal{A}}{\mathrm{min}}\,\, \frac{1}{2N}\lVert \mathbf{A} \circ \mathbf{C} \rVert_{1,1}  -\alpha \mathbf{1}^{\top}\mathrm{log}(\mathbf{A}\mathbf{1}) + {\beta}\lVert \mathbf{A}\rVert_{\mathrm{F}}^2,
    \label{GL-W}
    \end{align}
\end{shrinkfix}
where $\lVert \cdot \rVert_{1,1}$ is the element-wise $\ell_1$ norm of a matrix. Besides, $ \mathbf{C}\in \mathbb{R}^{d\times d}$ is a pairwise distance matrix defined as 
\begin{shrinkfix}
\begin{align}
       \mathbf{C}{[ij]} = \lVert\widetilde{\mathbf{x}}_i - \widetilde{\mathbf{x}}_j\rVert_2^2,
    \label{definition-z}
\end{align}
\end{shrinkfix}
where $\widetilde{\mathbf{x}}_i\in\mathbb{R}^{N}$ is the $i$-th row vector of $\mathbf{X}$. Similarly, $\mathcal{A}$ is the set  containing all adjacency matrices,
\begin{shrinkfix}
\begin{align}
        \mathcal{A} = \left\{ \mathbf{A} : \mathbf{A}\in \mathbb{S}^{d\times d}, \mathbf{A}\geq 0,  \mathrm{diag}(\mathbf{A}) = \mathbf{0}\right\},
    \label{GL-2}
\end{align}
\end{shrinkfix}
where $\mathbf{A}\geq 0$ means that all elements of $\mathbf{A}$ are non-negative. By the definition of $\mathcal{A}$, the number of free variables of $\mathbf{A}$ is $p := \frac{d(d-1)}{2}$ \cite{kalofolias2016learn}.  For simplicity, we define a vector $\mathbf{w}\in \mathbb{R}^p$ whose elements are the upper triangle variables of $\mathbf{A}$. Then, problem \eqref{GL-W} can be rewritten into a vector form as
\begin{shrinkfix}
\begin{align}
 \underset{\mathbf{w}\geq 0}{\mathrm{min}}\,\,\frac{1}{N} \mathbf{z}^{\top}\mathbf{w} { -\alpha\mathbf{1}^{\top}\mathrm{log}(\mathbf{S}\mathbf{w}) + 2\beta\lVert \mathbf{w} \rVert_2^2} ,
    \label{reformulation-of-GL-W}
\end{align}
\end{shrinkfix}
where $\mathbf{S}$ is a linear operator satisfying $\mathbf{S}\mathbf{w} = \mathbf{A}\mathbf{1}$, and $\mathbf{z}$ is the vector form of the upper triangle elements of $\mathbf{C}$\footnote{The pairwise distance vector  $\mathbf{z}$ is calculated from $\mathbf{x}_1,\dots, \mathbf{x}_N$. Thus, $\mathbf{z}$ is also referred to as observation data in the following sections.}. 

\begin{figure}[t] 
    \centering
       \includegraphics[width=0.95\linewidth ]{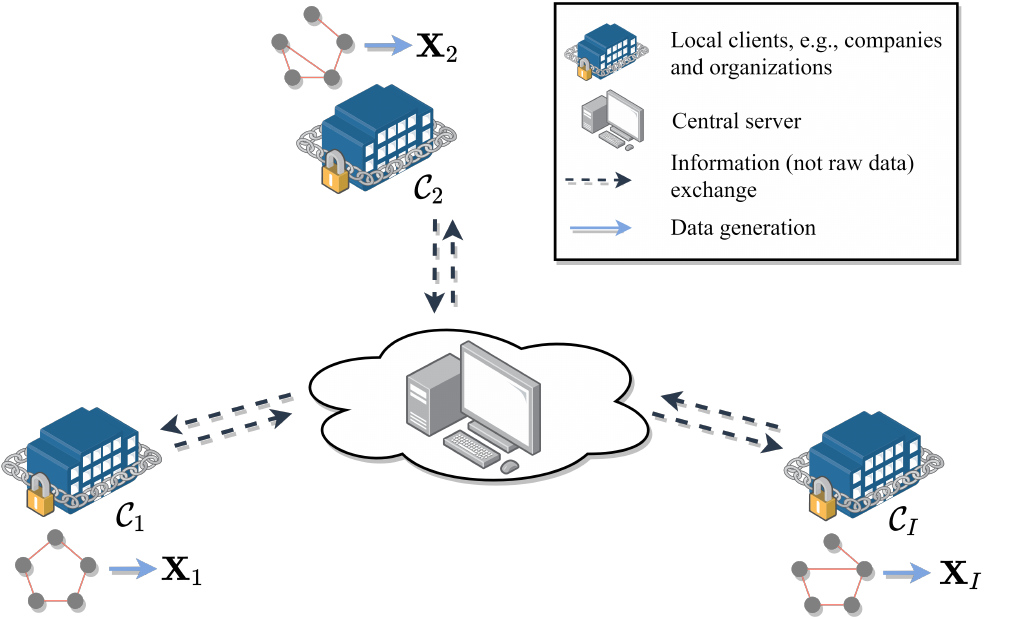}
    	\caption{The illustration of the target scenario. The clients $\mathcal{C}_1,\dots\mathcal{C}_I$ store graphs signals $\mathbf{X}_1,\dots,\mathbf{X}_I$ generated from $I$ distinct but related graphs. The data are not allowed to leave their clients, but all clients can commute with a central server.
    	}
    	\label{Fig-paradigm}
\end{figure}

\subsection{Problem Statement}
As shown in Fig.\ref{Fig-paradigm}, suppose there are $I$ clients $\mathcal{C}_1,\dots,\mathcal{C}_I$, and the $i$-th client stores signals $\mathbf{X}_i \in \mathbb{R}^{d\times N_i}$ generated from the graph $\mathcal{G}_i$,  where $N_i$ is the size of the $i$-th dataset. All graphs are defined over the same node set, and the data are prohibited from leaving the client where they are stored due to factors such as privacy concerns. However, the clients can exchange information with a central server. We assume that (\romannumeral1) graph signals $\mathbf{X}_i$ are smooth over  $\mathcal{G}_i$, and (\romannumeral2)  $\mathcal{G}_1,\dots,  \mathcal{G}_I$ may be heterogeneous, meaning that the corresponding graph signals could be non-IID. Our goal is to infer the graphs from  $\mathbf{X}_1,\dots, \mathbf{X}_I$ in the case of data silos.

\section{Model Formulation}
\label{sec:formulation-1}
In this section, we first propose an auto-weighted MGL model to learn graphs in the target scenario, and then analyze its estimation error bound.

\subsection{Basic Formulation}
We assume that $I$ local graphs, which are denoted as $\mathbf{w}_1,\dots, \mathbf{w}_I \in\mathbb{R}^p$, have some common structures named consensus graphs $\mathbf{w}_{\mathrm{con}}\in \mathbb{R}^p$. 
%This assumption is common in the real world, such as gene regulatory networks of the cells from the same tissue \cite{karaaslanli2021multiview}, and networks of brain functional connections between normally functioning brain regions in autistic and non-autistic individuals \cite{pu2021learning}. 
Intuitively, the consensus graph should be close to local graphs since they share common edges. Furthermore, the consensus graph should also be sparse to remove redundant noisy edges  \cite{hu2020multi}. Thus, we learn $I$ local graphs and the consensus graph jointly via
\begin{shrinkfix}
\begin{align}
\underset{\mathbf{w}_i, \mathbf{w}_{\mathrm{con}} \in \mathcal{W} }{\min} & \sum_{i=1}^I \underbrace{\frac{1}{N_i}  \mathbf{z}_{i}^{\top}\mathbf{w}_i - \alpha\mathbf{1}^{\top} \log\left(\mathbf{S}\mathbf{w}_i\right) + 2\beta\lVert \mathbf{w}_i\rVert_2^2}_{{g}_i(\mathbf{w}_i)} \notag\\ 
&+ {\lambda\nu\sum_{i=1}^I \lVert \mathbf{w}_i - \mathbf{w}_{\mathrm{con}}\rVert_2 + \lambda \lVert \mathbf{w}_{\mathrm{con}} \rVert_1 } \notag\\
=\underset{\mathbf{w}_i, \mathbf{w}_{\mathrm{con}} \in \mathcal{W} }{\min}& G(\mathbf{W}) + \lambda R (\mathbf{W})
    \label{sec:con-eq-1}
\end{align}
\end{shrinkfix}
where $\mathbf{z}_{i}$ is the observed data in client $\mathcal{C}_i$, $\mathcal{W} := \left\{\mathbf{w}: \mathbf{w}\geq 0  \right\}$, $\mathbf{W} = [\mathbf{w}_1,\dots,\mathbf{w}_I, \mathbf{w}_{\mathrm{con}}] \in \mathbb{R}^{p\times {(I+1)}}$, $G(\mathbf{W}): = \sum_{i=1}^I g_i(\mathbf{w}_i)$, and $R (\mathbf{W}) := \nu\sum_{i=1}^I \lVert \mathbf{w}_i - \mathbf{w}_{\mathrm{con}}\rVert_2 + \lVert \mathbf{w}_{\mathrm{con}} \rVert_1$. Our model consists of two parts. The first part $G(\mathbf{W})$ represents that local clients utilize the traditional smoothness-based model \eqref{reformulation-of-GL-W} to learn graphs $\mathbf{w}_1,\dots,\mathbf{w}_I$ from locally stored data. The second term $R(\mathbf{W})$ is the regularizer that topologically connects local graphs via the consensus graph. Specifically, the first term $\nu \sum_{i=1}^I\lVert \mathbf{w}_i - \mathbf{w}_{\mathrm{con}}\rVert_2$ of $R(\mathbf{W})$ measures difference between local graphs and the consensus graph. We add the $\ell_1$ norm term because $\mathbf{w}_{\mathrm{con}}$ is expected to be sparse. The constants  $\nu$  and $\lambda$ are predefined parameters. The parameters of problem \eqref{sec:con-eq-1} are $\alpha, \beta, \nu$, and $\lambda$. However, as stated in \cite{kalofolias2016learn}, tuning the importance of the log-degree term w.r.t. the other graph terms has a scaling effect. We can fix $\alpha$ and search for the other parameters. Therefore, the free parameters of our model are $\beta$, $\nu$, and $\lambda$.

The proposed model \eqref{sec:con-eq-1}  enjoys the following advantages. Firstly, our model provides a way to learn local personalized graphs $\mathbf{w}_1,\dots,\mathbf{w}_I$ jointly. Compared to learning graphs independently, our model utilizes the information from all datasets,  which could boost learning performance. Secondly, we learn a personalized graph for each client, which alleviates the bias of learning a single global graph using all data in the case of data heterogeneity. Lastly, unlike most PFL methods that only learn personalized local models \cite{tan2022towards}, our model learns a consensus graph reflecting global information.

\subsection{Auto-weighted Multiple Graph Learning}
\label{sec:auto-weighted}
At first glance, our model \eqref{sec:con-eq-1}  treats $I$ local graphs equally since they share the same weight in $\nu\sum_{i=1}^I \lVert \mathbf{w}_i -\mathbf{w}_{\mathrm{con}}\rVert_2$. This is unreasonable because the similarity between the consensus graph and different local graphs could vary widely. Fortunately, we will show that our model \eqref{sec:con-eq-1} can implicitly assign appropriate weights to local graphs, which we termed contribution weights, based on their similarity to $\mathbf{w}_{\mathrm{con}}$ via the inverse distance weighting schema \cite{nie2017self}.

Specifically, if we solve \eqref{sec:con-eq-1} by alternately updating $\mathbf{w}_1,\dots,\mathbf{w}_{I}$ and $\mathbf{w}_{\mathrm{con}}$\textemdash which is exactly how we solve it in Section \ref{sec:algorithm}\textemdash when we fix the consensus graph $\mathbf{w}_{\mathrm{con}}$, the sub-problem of updating $\mathbf{w}_i$ is  
\begin{shrinkfix}
\begin{align}
\underset{\mathbf{w}_{i}\in\mathcal{W} }{\min} \;\; g_i(\mathbf{w}_i) + \rho \lVert \mathbf{w}_i - \mathbf{w}_{\mathrm{con}}\rVert_2,
    \label{sec:con-eq-Lag-1-3}
\end{align}
\end{shrinkfix}
where we let $\rho = \lambda\nu$ for simplicity. We take the derivative of the Lagrange
function of \eqref{sec:con-eq-Lag-1-3} and set it to zero, which yields
\begin{shrinkfix}
\begin{align}
\rho \widetilde{\gamma}_i \frac{\partial \lVert \mathbf{w}_i - \mathbf{w}_{\mathrm{con}}\rVert_2^2}{\partial \mathbf{w}_i} + \frac{\partial g_i(\mathbf{w}_i)}{\partial \mathbf{w}_i} - \frac{\partial \bm{\theta}_i^{\top}\mathbf{w}_i}{\partial \mathbf{w}_i}= \mathbf{0},
    \label{sec:con-eq-Lag-1-4}
\end{align}
\end{shrinkfix}
where $\bm{\theta}_i\in\mathbb{R}^{p}$ is the Lagrange multiplier, and
\begin{shrinkfix}
\begin{align}
\widetilde{\gamma}_i = \frac{1}{2\lVert \mathbf{w}_i - \mathbf{w}_{\mathrm{con}} \rVert_2}.
    \label{sec:con-eq-Lag-1-1}
\end{align}
\end{shrinkfix}
Note that \eqref{sec:con-eq-Lag-1-1} depends on $\mathbf{w}_i$,  meaning that $\widetilde{\gamma}_i$ and $\mathbf{w}_i$ are coupled with each other. However, if we set $\widetilde{\gamma}_i$ stationary, \eqref{sec:con-eq-Lag-1-4} is the solution to the following problem
\begin{shrinkfix}
\begin{align}
\underset{\mathbf{w}_{i}\in\mathcal{W} }{\min} \;\; g_i(\mathbf{w}_i) + \rho  \widetilde{\gamma}_i\lVert \mathbf{w}_i - \mathbf{w}_{\mathrm{con}}\rVert_2^2.
    \label{sec:con-eq-Lag-1-3-1}
\end{align}
\end{shrinkfix}
After solving  \eqref{sec:con-eq-Lag-1-3-1}, we can use the obtained $\mathbf{w}_i$ to update the weight $\widetilde{\gamma}_i$ via \eqref{sec:con-eq-Lag-1-1}. Therefore, we can alternately update $\mathbf{w}_i$ and  $\widetilde{\gamma}_i$ to solve \eqref{sec:con-eq-Lag-1-3}. Similarly, it is not difficult to check that we can update $\mathbf{w}_{\mathrm{con}}$ using the same strategy as updating $\mathbf{w}_i$.

Combing the alternative updates of $\mathbf{w}_i, \mathbf{w}_{\mathrm{con}}$, and $\widetilde{\gamma}_i$, the basic formulation \eqref{sec:con-eq-1} is rephrased as
\begin{shrinkfix}
\begin{align}
&\underset{\mathbf{w}_i, \mathbf{w}_{\mathrm{con}}}{\min} \sum_{i=1}^I {\frac{1}{N_i}  \mathbf{z}_{i}^{\top}\mathbf{w}_i - \alpha\mathbf{1}^{\top} \log\left(\mathbf{S}\mathbf{w}_i\right) + 2\beta\lVert \mathbf{w}_i\rVert_2^2} \notag\\ 
&\;\;\;\;\;\;\;\; \;\;+{\frac{\rho}{2}\sum_{i=1}^I  {\gamma}_i\lVert \mathbf{w}_i - \mathbf{w}_{\mathrm{con}}\rVert_2^2 + \lambda \lVert \mathbf{w}_{\mathrm{con}} \rVert_1 } \notag\\
& \mathrm{s.t.}\; \mathbf{w}_i, \mathbf{w}_{\mathrm{con}} \in\mathcal{W}, \gamma_i = \frac{1}{\lVert\mathbf{w}_i -\mathbf{w}_{\mathrm{con}}  \rVert_2}.
    \label{sec:con-eq-1-rephrase}
\end{align}
\end{shrinkfix}
One merit of solving \eqref{sec:con-eq-1} via the reformulation \eqref{sec:con-eq-1-rephrase} is that it naturally produces a contribution weight ${\gamma}_i$ for the local graph $\mathbf{w}_i$. The weight value is determined by the similarity between $\mathbf{w}_i$ and $\mathbf{w}_{\mathrm{con}}$. For the $\mathbf{w}_i$  close to $\mathbf{w}_{\mathrm{con}}$, a larger ${\gamma}_i$ is assigned to the corresponding term,  increasing the contribution of the $i$-th local graph to the consensus graph. On the contrary, the local graph far from the consensus graph obtains a small ${\gamma}_i$. Thus, our model can implicitly and automatically assign appropriate weights to local clients based on their similarity to the consensus graph.

\subsection{Theoretical Analysis}
Let $\mathbf{W}^* = [\mathbf{w}^*_1,\dots,\mathbf{w}^*_I, \mathbf{w}^*_{\mathrm{con}}] \in \mathbb{R}^{p\times {(I+1)}}$ be the true graphs and $\widehat{\mathbf{W}} = [\widehat{\mathbf{w}}_1,\dots,\widehat{\mathbf{w}}_I, \widehat{\mathbf{w}}_{\mathrm{con}}] \in \mathbb{R}^{p\times {(I+1)}}$ be the estimated graphs of our model \eqref{sec:con-eq-1}. We aim to derive the estimation error bound of our proposed multiple graph estimator. Before conducting the analysis, we make the following assumptions.

\begin{assumption}
The observed signals, as well as the corresponding pairwise distance vectors $\mathbf{z}_i$ are bounded, i.e., there exists a constant $C_z$ such that  $\lVert\mathbf{z}_i\rVert_2 \leq C_z$ for $i=1,\dots,I$.
\label{assumption-0}
\end{assumption}

\begin{assumption}
Let $h(\mathbf{w} ) = - \alpha\mathbf{1}^{\top} \log\left(\mathbf{S}\mathbf{w}\right) + 2\beta\lVert \mathbf{w}\rVert_2^2$. The gradient of $h(\mathbf{w})$ at the real graph is bounded, i.e., there exists a constant $C_h$ such that $\left\lVert \nabla h(\mathbf{w}_i^*) \right\rVert_2 \leq C_h$ for $i=1,\dots,I$.
\label{assumption-1-1}
\end{assumption}
Assumption \ref{assumption-0} naturally holds in the real world since unbounded signals do not make sense. Assumption \ref{assumption-1-1} holds when the true graph has limited edge weights and no isolated nodes, which is common in the real world. Moreover, without loss of generality, we assume the data sizes of $I$
graphs are the same, i.e., $N_i = N$ for $ i=1,\dots,I$. The analysis can be easily generalized to the case where data sizes vary across different clients. We further suppose that the obtained data vector (pairwise distance vector) $\mathbf{z}_i$ can be written as  
\begin{shrinkfix}
\begin{align}
\mathbf{z}_i[j] = \mathbf{z}^*_i[j] + \mathbf{e}_i[j], \;j=1,\dots,p,
    \label{eq-theo-0-1}
\end{align}
\end{shrinkfix}
where $\mathbf{z}^*_i$ is the true data vector, and $\mathbf{e}_i$ is the error vector caused by factors such as noisy measurements. In our analysis, we assume that $\mathbf{e}_i\sim\mathcal{N}(\mathbf{0},\sigma_e^2\mathbf{I})$. Then, we present the estimation error bound of the proposed graph estimator as stated in the following theorem.

\begin{theorem}
Under Assumptions \ref{assumption-0} and \ref{assumption-1-1}, given $\delta$ and $ \nu >0$, let $\lambda$ satisfy 
\begin{shrinkfix}
\begin{align}
\lambda \geq \frac{C_z\sqrt{I}  + \sigma_e\sqrt{pI+ \delta}}{C_rN},
    \label{eq-lambda-selection}
\end{align}
\end{shrinkfix}
where $C_r:=\nu \sqrt{I\omega_{\max}(\mathbf{L}_m)}+ \sqrt{p}$, 
\begin{shrinkfix}
\begin{align}
\mathbf{L}_m \in \mathbb{R}^{(I+1)\times (I+1)}= 
\begin{bmatrix}
1 & 0 & \cdots & 0&-1\\
0 & 1 & \cdots & 0&-1\\
\vdots & \ddots &\cdots& 0 &-1 \\
0 & 0  & \cdots & 1 &-1\\
-1 & -1  & -1 & -1 & I
\end{bmatrix},
    \label{eq-theo-2}
\end{align}
\end{shrinkfix}
and $\omega_{\max}(\mathbf{L}_m)$ is the maximum  eigenvalue of $\mathbf{L}_m$. Then, we have probability at least $1 -\exp\left( -\frac{1}{2}\left(\delta - pI\log\left(1 + \frac{\delta}{pI} \right)\right)\right)$ such that 
\begin{shrinkfix}
\begin{align}
 \left\lVert  \widehat{\mathbf{W}} - \mathbf{W}^*\right\rVert_{\mathrm{F}}
 \leq  \frac{C_r \lambda }{\beta} + \frac{C_h \sqrt{I}}{2\beta}.
    \label{eq-theo-1}
\end{align}
\end{shrinkfix}
    \label{theo-estimation-error}
\end{theorem}
\begin{proof}
%The proof is mainly inspired by the derived from the work \cite{negahban2012unified} but with proper adjustments to our problem, which is not trivial. 
See Appendix \ref{appendix-0} for details.
\label{proof-estimation-error}
\end{proof}
Theorem \ref{theo-estimation-error} characterizes the estimation error bound of our proposed model w.r.t. some key factors, such as graph size $d$ (the number of the free variables $p$), data size $N$, the number of local graphs $I$, and measurement noise level $\sigma_e$. The theorem states that, if the weight $\lambda$ before the regularizer $R(\mathbf{W})$ is selected appropriately,  the estimation error of our model is bounded by the r.h.s. of \eqref{eq-theo-1} with high probability. The upper bound consists of two parts. The first part is related to data sizes. When $N$ goes to infinity, \eqref{eq-lambda-selection} indicates that $\lambda$ could be small enough that the first part decreases to zero. The second part is determined by the regularizer $h(\mathbf{w}^*)$ of the basic graph learning model \eqref{reformulation-of-GL-W}. This can be regarded as a systematic error since $h(\mathbf{w})$ is chosen using prior knowledge, and the error will not decrease as $N$ increases. We should mention that our model is flexible, and one can choose any $h(\mathbf{w})$ as the regularizer of local models to bring desired properties to the learned graphs. For the well-selected regularizer, the gradient of $h(\mathbf{w})$ at $\mathbf{w}^*$ will be bounded tightly, and we can obtain a small upper bound on the estimation error.

\subsection{Connections to  Existing MGL Models}
In the literature, there have been some works that learn multiple graphs based on the assumption of common structures. These works differ mainly in how they describe structural relationships among multiple graphs. To avoid notational confusion, we use different notations to interpret existing works. We let $\mathbf{K}\in\mathbb{R}^{d\times d}$ denote the common graph structures and $\mathbf{U}_{1},\dots, \mathbf{U}_{I}\in\mathbb{R}^{d\times d}$ represent the unique structures of $I$ local graphs. Under these notations, the graph regularizer $R(\mathbf{W})$ in our model \eqref{sec:con-eq-1} can be roughly summarized as (not exactly equivalent to) $\lVert \mathbf{K}\rVert_{1,1} + \nu \sum_{i=1}^I \lVert \mathbf{U}_i\rVert_{\mathrm{F}} $. Unlike our model, the work \cite{hara2013learning} defines the regularizer as $\lVert \mathbf{K}\rVert_{1,1} + \nu \lVert \mathbf{U} \rVert_{1,r}$, where $r\in [1, +\infty)$, $ \lVert \mathbf{U} \rVert_{1,r} = \sum_{i,j=1}^d \lVert \mathbf{U}[ij]\rVert_r$, and $\mathbf{U}[ij] = [\mathbf{U}_{1}[ij],\dots, \mathbf{U}_{I}[ij]]^{\top}\in \mathbb{R}^I$. The term $\lVert \mathbf{U} \rVert_{1,r}$ is a group lasso regularizer that can capture common sparsity structure among all local graphs. The second work \cite{lee2015joint} assumes that the common structure $\mathbf{K}$ is the average of all local graphs, i.e., $\mathbf{K} = \frac{1}{I}\sum_{i=1}^{I} \mathbf{A}_i, \mathbf{U}_i = \mathbf{A}_i - \mathbf{K}$. A regularizer is then designed as $ \lVert \mathbf{K}\rVert_{1,1} + \nu \sum_{i=1}^I \lVert \mathbf{U}_i\rVert_{1,1} $. Unlike our work, \cite{karaaslanli2021multiview} proposes a regularizer $\lVert \mathbf{K}\rVert_{1,1} + \nu \sum_{i=1}^I \lVert \mathbf{U}_i\rVert_{\mathrm{F}}^2 $. Moreover, the local graph learning model of \cite{karaaslanli2021multiview} is different from ours, and \cite{karaaslanli2021multiview} requires all graphs, including the consensus graph,  are represented by Laplacian matrices. Recently, \cite{karaaslanli2024multiview} designs two regularizers, i.e., $ \sum_{i=1}^I \lVert \mathbf{U}_i\rVert_{1,1} $ and the other similar to \cite{hara2013learning}.

In summary, our model differs from existing works in the following aspects. (\romannumeral1) The regularizer $R(\mathbf{W})$ is different from existing works, which can implicitly assign a contribution weight for each local graph. This is used for the first time in the multiple graph learning problem. (\romannumeral2) Our model and \cite{karaaslanli2021multiview} learn multiple graphs based on the smoothness assumption \eqref{GL-2}, while \cite{hara2013learning, lee2015joint} are based on the GGMs. (\romannumeral3) We further consider the problem of learning graphs in the case of data silos, which has not been explored in the existing GL-related works.

\section{Model Optimization}
\label{sec:algorithm}
In this section, we propose an algorithm with provable convergence analysis for solving \eqref{sec:con-eq-1}  in the target scenario. 

\subsection{The Proposed Algorithm}
\label{sec:alg-1}

According to the reformulation \eqref{sec:con-eq-1-rephrase}, our algorithm consists of two  steps: (\romannumeral1) updating $\mathbf{w}_i$ in the local client $\mathcal{C}_i$ and (\romannumeral2) updating $\bm{\gamma} = [\gamma_1,\dots, \gamma_I]^{\top}\in\mathbb{R}^{I}$ and $\mathbf{w}_{\mathrm{con}}$ in the central server. The complete flow is displayed in Algorithm  \ref{alg:FGL}.

\textbf{Updating $\mathbf{w}_i$ in the local client $\mathcal{C}_i$:} This update corresponds to lines 3-11 in Algorithm \ref{alg:FGL}. In the $t$-th communication round,  $\mathcal{C}_i$ first receives $\mathbf{w}^{(t)}_{\mathrm{con}}$ and $\gamma_i^{(t)}$ from the central server. Let $f_i^{(t)}(\mathbf{w}_i,\mathbf{w}_{\mathrm{con}}) := g_i(\mathbf{w}_i)+ \frac{\rho\gamma_i^{(t)}}{2}\lVert \mathbf{w}_i - \mathbf{w}_{\mathrm{con}}\lVert_2^2$ \footnote{$f_i^{(t)}(\mathbf{w}_i,\mathbf{w}_{\mathrm{con}}^{(t)})$ is a function of $\mathbf{w}_i$ with fixed $\mathbf{w}_{\mathrm{con}}^{(t)}$. Similarly, $f_i^{(t)}(\mathbf{w}_i^{(t+1)},\mathbf{w}_{\mathrm{con}})$ is a function of $\mathbf{w}_{\mathrm{con}}$ with fixed $\mathbf{w}_i^{(t+1)}$. }, and the sub-problem of $\mathcal{C}_i$ becomes
%\footnote{In the sequel, $f_i^{(t)}(\mathbf{w}_i,\mathbf{w}_{\mathrm{con}}^{(t)})$ is the function of variable $\mathbf{w}_i$ with fixed $\mathbf{w}_{\mathrm{con}}^{(t)}$ and $\gamma^{(t)}_i$, $f_i(\mathbf{w}_i^{(t)},\mathbf{w}_{\mathrm{con}})$ is the function of variable $\mathbf{w}_{\mathrm{con}}$ with fixed $\mathbf{w}_{i}^{(t)}$ and  and $\gamma^{(t)}_i$.
%and  $f_i(\mathbf{w}_i,\mathbf{w}_{\mathrm{con}})$ is  the function of both $\mathbf{w}_i$ and $\mathbf{w}_{\mathrm{con}}$.}
\begin{shrinkfix}
\begin{align}
 \mathbf{w}_i^{(t+1)} = 
 &\underset{\mathbf{w}_i\in \mathcal{W}}{\mathrm{argmin}}\,\, f_i^{(t)}(\mathbf{w}_i,\mathbf{w}_{\mathrm{con}}^{(t)})\notag\\
 =&\underset{\mathbf{w}_i\in \mathcal{W}}{\mathrm{argmin}} \,\frac{1}{N_i}  \mathbf{z}_{i}^{\top}\mathbf{w}_i - \alpha\mathbf{1}^{\top} \log\left(\mathbf{S}\mathbf{w}_i + \zeta\mathbf{1}\right) \notag\\
 &+ 2\beta\lVert \mathbf{w}_i\rVert_2^2+ \frac{\rho\gamma_i^{(t)}}{2}\lVert \mathbf{w}_i - \mathbf{w}^{(t)}_{\mathrm{con}}\rVert_2^2,
\label{sec:alg-eq-1}
 \end{align}
\end{shrinkfix}
where $\zeta$ is a small enough constant to avoid zero node degree. We use the accelerated projected gradient descent algorithm to update $\mathbf{w}_i$.  At the beginning of local updates,  $\mathcal{C}_i$ first initializes $\mathbf{w}_i^{(t,0)} = \mathbf{w}_i^{(t-1,K_i^{(t-1)})}$ and $\mathbf{w}_i^{(t,-1)} = \mathbf{w}_i^{(t-1,K_i^{(t-1)}-1)}$, where $K_i^{(t)}$ is the number of local loops of $\mathcal{C}_i$ in the $t$-th outer iteration. Besides, $\mathbf{w}_i^{(t,k)}$ represents the updated graph of $\mathcal{C}_i$ in the $t$-th outer iteration  and the $k$-th local loop. We then  update the $i$-th local graph  $K_i^{(t)}$ times by 
\begin{shrinkfix}
\begin{align}
&\mathbf{w}_{i,\mathrm{ex}}^{(t,k)} = \mathbf{w}_{i}^{(t,k)} + \xi\left(\mathbf{w}_{i}^{(t,k)} - \mathbf{w}_{i}^{(t,k-1)}\right)\label{sec:alg-eq-2-0}\\ 
&\breve{\mathbf{w}}_{i}^{(t,k+1)} = \mathbf{w}_{i,\mathrm{ex}}^{(t,k)} - \eta_w\bigg{(} \nabla g_i(\mathbf{w}_{i,\mathrm{ex}}^{(t,k)}) \notag\\
&\;\;\;\;\;\;\;\;\;\;\;\;\;\;\;\;\;\;\;\;\;\;\;\;\;\;\;\;\;\;\;\;\;\;\;\;\;\;+ \rho\gamma_i^{(t)}\left(\mathbf{w}_{i,\mathrm{ex}}^{(t,k)} - \mathbf{w}_{\mathrm{con}}^{(t)}\right)\bigg{)}\label{sec:alg-eq-2-1} \\
&\mathbf{w}_{i}^{(t,k+1)} = \mathrm{Proj}_{\mathcal{W}} \left(\breve{\mathbf{w}}_{i}^{(t,k+1)}\right),
\label{sec:alg-eq-2-2}
\end{align}
\end{shrinkfix}
where $\xi \in [0,1)$ is an momentum weight. In \eqref{sec:alg-eq-2-1},  $\nabla g_i(\mathbf{w})$ is calculated as $\frac{1}{N_i}\mathbf{z}_{i} - \alpha\mathbf{S}^{\top}\left(\frac{1}{\mathbf{S}\mathbf{w} + \zeta \mathbf{1}}\right) + 4 \beta\mathbf{w}$. Moreover, $\eta_w$ is the stepsize, the choice of which will be discussed in the next subsection. The operator $\mathrm{Proj}_{\mathcal{W}} \left(\cdot\right)$ means projecting variables into  ${\mathcal{W}}$.  When local updates finish, $\mathcal{C}_i$  sends $\mathbf{w}_i^{(t+1)} = \mathbf{w}_i^{(t,K_i^{(t)})}$ to the central server. Note that the local update of $\mathbf{w}_i$ is inexact in one communication round since we run \eqref{sec:alg-eq-2-0}-\eqref{sec:alg-eq-2-2} for $K_i^{(t)}$ times without convergence. We use inexact updates because we aim to update all local graphs synchronously. Due to system heterogeneity, the time required for each client to update $\mathbf{w}_i$ until convergence may vary widely. It takes longer if the central server waits for all clients to update their graphs until convergence.

\textbf{Updating $\mathbf{w}_{\mathrm{con}}$ and $\bm{\gamma}$ in the central server:} The updates correspond to lines 12-14 in Algorithm \ref{alg:FGL}. After the central server receives all the updated local graphs, $\mathbf{w}_1^{(t+1)},\dots,\mathbf{w}_I^{(t+1)}$, we  update $\mathbf{w}_{\mathrm{con}}^{(t+1)}$ by solving the following problem 
\begin{shrinkfix}
\begin{align}
\mathbf{w}_{\mathrm{con}}^{(t+1)} &= \underset{\mathbf{w}_{\mathrm{con}}\in\mathcal{W}}{\mathrm{argmin}}\,\, \sum_{i=1}^{I} f_i^{(t)}(\mathbf{w}_i^{(t+1)}, \mathbf{w}_{\mathrm{con}})  + \lambda\lVert \mathbf{w}_{\mathrm{con}}\rVert_1\notag\\
&=\underset{\mathbf{w}_{\mathrm{con}}\in\mathcal{W}}{\mathrm{argmin}} \,\sum_{i=1}^{I} \frac{\rho\gamma_i^{(t)}}{2}\lVert \mathbf{w}_i^{(t+1)} - \mathbf{w}_{\mathrm{con}}\rVert_2^2 + \lambda\lVert \mathbf{w}_{\mathrm{con}}\rVert_1.
\label{sec:alg-eq-3}
\end{align}
\end{shrinkfix}
The problem is equivalent to 
\begin{shrinkfix}
\begin{align}
 \mathbf{w}_{\mathrm{con}}^{(t+1)} =& \underset{\mathbf{w}_{\mathrm{con}}\in\mathcal{W}}{\mathrm{argmin}} \, \frac{1}{2}\left\lVert \mathbf{w}_{\mathrm{con}} - \frac{\sum_{i=1}^I \gamma_i^{(t)}\mathbf{w}_i^{(t+1)}}{\sum_{i=1}^I \gamma_i^{(t)}}\right\rVert_2^2 \notag\\
 &\;\;\;\;\;\;\;\;\;\;\;+
\frac{\lambda}{\rho\sum_{i=1}^I\gamma_i^{(t)}}\lVert \mathbf{w}_{\mathrm{con}}\rVert_1.
\label{sec:alg-eq-4}
\end{align}
\end{shrinkfix}
By defining $C_{\gamma}^{(t)} := \sum_{i=1}^I \gamma_i^{(t)}$  and  $\mu^{(t)} := \frac{\lambda}{C_{\gamma}^{(t)}\rho}$, we obtain
\begin{shrinkfix}
\begin{align}
\mathbf{w}_{\mathrm{con}}^{(t+1)} = \mathrm{prox}_{\mu^{(t)}\lVert \cdot\rVert_1}\left(\frac{\sum_{i=1}^I \gamma_i^{(t)}\mathbf{w}_i^{(t+1)}}{C_{\gamma}^{(t)}}\right),
\label{sec:alg-eq-45}
\end{align}
\end{shrinkfix}
where $\mathrm{prox}_{\mu^{(t)}\lVert \cdot\rVert_1}(\cdot)$ is the proximal operator of $\ell_1$ norm. It is observed from \eqref{sec:alg-eq-45} that $\mathbf{w}_{\mathrm{con}}^{(t+1)} \in \mathcal{W}$ since it is a  combination of $\mathbf{w}_i^{(t+1)}$ with positive weights, and the proximal operator will not move the vector out of $\mathcal{W}$.

After obtaining $\mathbf{w}_{\mathrm{con}}^{(t+1)}$, we  update  $\gamma_i^{(t+1)}$ as\footnote{
To avoid dividing by zero, we can update $\gamma_i^{(t+1)}$ as $\frac{1}{\left\lVert \mathbf{w}_{i}^{(t+1)} -  \mathbf{w}_{\mathrm{con}}^{(t+1)}\right\rVert_2 + \epsilon_{\gamma}}$, where $\epsilon_{\gamma}$ is a small enough constant.
}
\begin{shrinkfix}
\begin{align}
\gamma_i^{(t+1)} = \frac{1}{\left\lVert \mathbf{w}_{i}^{(t+1)} -  \mathbf{w}_{\mathrm{con}}^{(t+1)}\right\rVert_2}, \,\, \mathrm{for}\,\, i = 1,\dots,I.
\label{sec:alg-eq-5-new}
\end{align}
\end{shrinkfix}
Finally, we send $\mathbf{w}_{\mathrm{con}}^{(t+1)}$ and $\gamma_i^{(t+1)}$ back to the client $\mathcal{C}_i$.

\begin{algorithm}[t] 
\caption{The algorithm for solving \eqref{sec:con-eq-1}} 
\begin{algorithmic}[1] %这个1 表示每一行都显示数字
\REQUIRE  %算法的输入参数：Input
$\alpha$, $\beta$, $\nu$, $\xi$, $\lambda$, and  signals $\mathbf{X}_1,\dots,\mathbf{X}_I$ \\
%\ENSURE ~~\\ %算法的输出：Output The learned graph $\mathbf{w}_1$, \ldots ,$\mathbf{w}_T$\\
\STATE  \textbf{Initialize}  $\gamma_i^{(0)} = 1/I$, $\mathbf{w}_{\mathrm{con}}^{(0)} = \mathbf{w}_{i}^{(0)}= \mathbf{w}_i^{(-1,0)} = \mathbf{w}_i^{(-1,-1)}$ for $i =1,\dots,I$, and let $K_i^{(-1)} = 0$\\
\FOR{$t = 0,\dots, T-1$}
\STATE {\color{red}\emph{/\,/\,\,Update $\mathbf{w}_1,\dots,\mathbf{w}_I$ in parallel in  local clients}} 
\FOR{$i = 1,\dots, I$ in parallel}
\STATE Receive $\gamma_i^{(t)}$ and  $\mathbf{w}^{(t)}_{\mathrm{con}}$ from the central server
\STATE Initialize  $\mathbf{w}_i^{(t,0)} = \mathbf{w}_i^{(t-1,K_i^{(t-1)})}$ and $\mathbf{w}_i^{(t,-1)} = \mathbf{w}_i^{(t-1,K_i^{(t-1)}-1)}$
\FOR{$k = 0,\dots, K_i^{(t)} - 1$}
\STATE Update $\mathbf{w}_{i}^{(t, k+1)}$ using \eqref{sec:alg-eq-2-0}-\eqref{sec:alg-eq-2-2} 
\ENDFOR
\STATE Let $\mathbf{w}_i^{(t+1)} = \mathbf{w}_i^{(t,K_i^{(t)})}$ and send it to central server
\ENDFOR

\STATE {\color{red}\emph{/\,/\,\, Update $\mathbf{w}_{\mathrm{con}}$ and $\bm{\gamma}$ in central server}}
\STATE Update $\mathbf{w}_{\mathrm{con}}^{(t+1)}$ using \eqref{sec:alg-eq-45}

\STATE Update  ${\gamma}_i^{(t+1)}$ using \eqref{sec:alg-eq-5-new}, for $i = 1,\dots,I$

\STATE Send $\mathbf{w}_{\mathrm{con}}^{(t+1)}$ and $\gamma_i^{(t+1)}$ to the $i$-th client $\mathcal{C}_i$

\ENDFOR
\RETURN $\mathbf{w}_1^{(T)}$, \ldots ,$\mathbf{w}_I^{(T)}$ and $\mathbf{w}_{\mathrm{con}}^{(T)}$ %算法的返回值
\end{algorithmic}
\label{alg:FGL} 
\end{algorithm}

\subsection{Convergence Analysis}
Before proceeding with the convergence analysis, let us first discuss the properties of the objective function.

\begin{proposition}
The objective function $f_i^{(t)}(\mathbf{w}_i,\mathbf{w}_{\mathrm{con}}^{(t)})$ is $L_i^{(t)}$-Lipschitz smooth w.r.t. $\mathbf{w}_i$ on $\widetilde{\mathcal{W}}$, where $\widetilde{\mathcal{W}} = \mathrm{conv}[\cup_{0\leq\xi\leq1} \{\mathbf{y} + \xi (\mathbf{y} - \mathbf{y}^{\prime}) ||\; \mathbf{y},\mathbf{y}^{\prime}\in\mathcal{W}\}]$ is  the feasible set extended by the momentum update, and $L_i^{(t)} := 4\beta + {2\alpha(d-1)}/{\zeta^2} + \rho\gamma_i^{(t)}$. Moreover,  $f_i^{(t)}(\mathbf{w}_i,\mathbf{w}_{\mathrm{con}}^{(t)})$ is $S_i^{(t)}$-strongly convex w.r.t. $\mathbf{w}_{i}$, where $S_i^{(t)} = 4\beta + \rho \gamma_i^{(t)} $.
\label{fact-1}
\end{proposition}

\begin{proof}
We place the proof in Appendix  \ref{appendix-1}.
\label{proof-proposition-1}
\end{proof}

Proposition \ref{fact-1} points out that the objective function $f_i^{(t)}(\mathbf{w}_i,\mathbf{w}_{\mathrm{con}}^{(t)})$  enjoys some nice properties, i.e., it is $L_i^{(t)}$-Lipschitz smooth and $S_i^{(t)}$-strongly convex, which are essential to the convergence of  accelerated gradient descent algorithms \cite{nesterov2013introductory}. Next, we further make the following assumptions.

\begin{assumption}
The feasible set is convex and compact.
\label{assumption-2}
\end{assumption}

\begin{assumption}
The (constant) stepsize satisfies  $\eta_w \leq 1/L_{max}$, where $L_{max}$ is the maximum value of $L_i^{(t)}$ for all $i$ and $t$.
\label{assumption-3}
\end{assumption}

Assumption \ref{assumption-2} is common in the constrained optimization algorithms. Assumption \ref{assumption-3} requires the (constant) stepsize to be bounded. Based on these assumptions, we conduct convergence analysis of our proposed algorithm as follows.

\begin{theorem}
Based on Assumptions \ref{assumption-2} and  \ref{assumption-3}, in each communication round of Algorithm \ref{alg:FGL}, the updated $\mathbf{w}_{\mathrm{con}}$ and $\mathbf{w}_i$  monotonically decrease the objective function of the proposed model \eqref{sec:con-eq-1} until the solution converges to a local optimum of the problem \eqref{sec:con-eq-1}.
\label{theorem-convergence}
\end{theorem}

\begin{proof}
See Appendix \ref{appendix-2} for details.
\end{proof}

The theorem reveals that as communication round $t$ increases, the proposed algorithm will converge to a local optimum of problem \eqref{sec:con-eq-1}. In Section \ref{sec:exper-convergence}, we will experimentally test the convergence of the proposed algorithm.

\subsection{Privacy Analysis}
In our algorithm, each client $\mathcal{C}_i$ uses its private data $\mathbf{X}_i$  to update the local graph $\mathbf{w}_i$ by solving \eqref{sec:alg-eq-1}. After local updates, clients send model updates instead of raw data to the central server to update the consensus graph and weights. Following the paradigm of FL,  all data is kept locally without any leakage during execution, which protects data privacy to a certain extent. Therefore, the proposed algorithm can learn graphs in the target scenario. However, inference attacks \cite{shokri2017membership} and model inversion attacks \cite{al2016reconstruction} show that the updates sent by local clients may still reveal private information. To further prevent information leakage, some popular techniques such as differential privacy \cite{dwork2006calibrating} may be helpful. This is beyond the scope of this study and will be left to future work.

\subsection{Some Discussions} 
\textbf{Differences from distributed learning algorithms}: Our algorithm may look similar to distributed learning algorithms at first glance, but there are actually fundamental differences. (\romannumeral1) Our algorithm falls into the emerging field of federated learning, which aims to train models on siloed datasets under privacy constraints, while distributed learning aims to employ multiple clients (computing nodes) to  accelerate training on a large but “flat” dataset \cite{kairouz2021advances}. (\romannumeral2) In distributed learning, data are centrally stored and can be shuffled and balanced across clients. Thus, data of different clients are IID. In contrast, in federated learning, data are generated locally and may be non-IID. In our problem, the data are generated from heterogeneous graphs and are hence non-IID. We propose to learn personalized graphs to alleviate the impact of non-IID data. (\romannumeral3) Distributed learning divides the dataset into multiple subsets uniformly and randomly, and  data sizes across different computing nodes are close. However, in federated learning,  all data are generated from local clients, which may be geographically and functionally diverse organizations, bringing  widely varying data sizes of different clients.

\textbf{Differences from existing FL algorithms}:  On the other hand, our algorithm differs from existing (personalized) FL algorithms in the following ways due to the specificity of our problem. (\romannumeral1) Unlike the algorithms of cross-device FL \cite{t2020personalized, mcmahan2017communication}, our algorithm learns graphs in a cross-silo FL scenario. Specifically, all clients, rather than a subset of clients, participate in local updates in each communication round. In addition, local clients utilize all stored data to update graphs instead of sampling a batch of data like many cross-device FL algorithms \cite{t2020personalized, mcmahan2017communication}. (\romannumeral2) The optimization problems corresponding to local and global updates are customized. For example, our algorithm updates the consensus graph by solving a $\ell_1$ norm regularized quadratic optimization problem, whereas the global update of existing algorithms is a simple aggregation of local models \cite{t2020personalized, smith2017federated, marfoq2021federated}. (\romannumeral3) Finally, compared to decentralized PFL algorithms \cite{bellet2018personalized, chen2022personalized}, our algorithm follows the ``central server-clients'' schema.

\section{Experiments}
\label{sec:Experiments}
In this section, we will test our proposed framework using both synthetic data and real-world data. First, some experimental setups are introduced.
\subsection{Experimental Setups}

\begin{table*}[t] 
\tabcolsep=0.5em
	\centering
 \begin{threeparttable}
	\caption{Performance of varying data sizes.}
	{\small
	\begin{tabular}{c|cccc|cccc|ccccc}
	\Xhline{1.2pt}
	
	 & \multicolumn{4}{c|}{$N = 20$} & \multicolumn{4}{c|}{$N = 50$}&\multicolumn{4}{c}{$N = 100$}\\

	\cline{2-13}
	 & $\mathrm{Precision}\uparrow$  &$\mathrm{Recall}\uparrow$  &$\mathrm{FS}\uparrow$  & $\mathrm{RE}\downarrow$  & $\mathrm{Precision}\uparrow $ &$\mathrm{Recall}\uparrow$& $\mathrm{FS}\uparrow$ & $\mathrm{RE}\downarrow$ & $\mathrm{Precision}\uparrow$  &$\mathrm{Recall}\uparrow$ & $\mathrm{FS}\uparrow$ & $\mathrm{RE}\downarrow$ \\

	\hline
        IGL                
        &{0.594} &0.469  &0.521 &0.915 
        &0.692 &0.620 & 0.651 &0.778 
        &0.776 &0.739 & 0.755 &0.698 \\	
    
     FedAvg       
     &0.434 &0.759  &0.552 &0.802  
     &0.464 &\textbf{0.792} & 0.584 &0.768 
     &0.519 &0.789 & 0.624 &0.744 \\

     FedProx      
     &0.423 &\textbf{0.767}  &0.544 &\textbf{0.765} 
     &0.523 &0.728 & 0.606 &0.718 
     &0.623 &0.703 & 0.655 &0.690 \\

     MRMTL      
     &0.522 &0.551  &0.547 &0.881  
     &0.647 &0.688 & 0.663 &0.756 
     &0.738 &0.789 & 0.760 &0.680 \\

     Ditto      
     &0.582 &0.547  &0.557 &0.884  
     &0.662 &0.671 & 0.663 &0.775 
     &0.732 &0.780 & 0.750 &0.696 \\     
    
    PPGL-L        
    &0.547 &0.579  &0.561 &0.911  
    &0.625 &0.760 & 0.678 &0.731 
    &0.698 &\textbf{0.870} & 0.774 &\textbf{0.619} \\
  
     PPGL-C               
     &\textbf{0.636} &0.516  &\textbf{0.566} &0.874  
     &\textbf{0.781} &0.656 & \textbf{0.710} &\textbf{0.701}
     &\textbf{0.881} &0.736 & \textbf{0.800} &0.627\\

	\Xhline{1.2pt}
	\end{tabular} 
	}
          \begin{tablenotes}  
        \footnotesize {         
        \item[1] In this paper, $\uparrow$ indicates that larger values are better, and $\downarrow$ indicates that smaller values are better.}   

      \end{tablenotes} 
       \label{table-datasize}
\end{threeparttable}

\end{table*}

\begin{table*}[t]
\tabcolsep=0.5em
	\centering
	\caption{Performance of varying data heterogeneity.}
	{\small
	\begin{tabular}{c|cccc|cccc|ccccc}
	\Xhline{1.2pt}
	
	 & \multicolumn{4}{c|}{$q = 0.3$} & \multicolumn{4}{c|}{$q = 0.6$}&\multicolumn{4}{c}{$q = 0.9$}\\

	\cline{2-13}
	 & $\mathrm{Precision}\uparrow$  &$\mathrm{Recall}\uparrow$  &$\mathrm{FS}\uparrow$  & $\mathrm{RE}\downarrow$  & $\mathrm{Precision}\uparrow $ &$\mathrm{Recall}\uparrow$& $\mathrm{FS}\uparrow$ & $\mathrm{RE}\downarrow$ & $\mathrm{Precision}\uparrow$  &$\mathrm{Recall}\uparrow$ & $\mathrm{FS}\uparrow$ & $\mathrm{RE}\downarrow$ \\

	\hline
        IGL               
        &\textbf{0.701} &0.593  &0.640 &0.783 
        &\textbf{0.736} &0.645 & 0.685 &0.758 
        &0.822 &0.706 & 0.759 &0.717 \\	
    
     FedAvg       
     &0.385 &\textbf{0.841}  &0.527 &0.835  
     &0.460 &\textbf{0.867} & 0.599 &0.783 
     &0.811 &0.922 & 0.862 &0.672 \\

     FedProx      
     &0.437 &0.720  &0.543 &0.770
     &0.569 &0.750 & 0.645 &0.716 
     &0.874 &0.863 & 0.868 &0.641 \\

     MRMTL      
     &0.662 &0.691  &0.674 &0.761 
     &0.690 &0.730 & 0.707 &0.753 
     &0.816 &0.794 & 0.807 &0.663 \\

     Ditto      
     &0.667 &0.675  &0.668 &0.765  
     &0.698 &0.714 & 0.704 &0.739 
     &0.821 &0.783 & 0.804 &0.667 \\

    PPGL-L        
    &0.637 &0.736  &\textbf{0.679} &\textbf{0.740} 
    &0.661 &0.776& \textbf{0.715} &\textbf{0.700}
    &0.800 &0.812 & 0.805 &0.635 \\
  
     PPGL-C               
     &0.462 &0.516  &0.487 &0.874 
     &0.653 &0.726 & 0.687 &0.752 
     &\textbf{0.825} &\textbf{0.927} & \textbf{0.873} &\textbf{0.634} \\

	\Xhline{1.2pt}
	\end{tabular} 
	}
 \label{table-heter}

\end{table*}

\textit{1) Graph generation: } We first generate Gaussian radial basis function (RBF) graphs $\mathcal{G}_0$ by following the method in \cite{dong2016learning}. Specifically, we generate 20 vertices whose coordinates are randomly in a unit square. Edge weights are calculated using $\mathrm{exp}(-\mathrm{dist}(i,j)^2/2\sigma^2_r)$, where $\mathrm{dist}(i,j)$ is the distance between vertices $i$ and $j$,  and $\sigma_r = 0.5$ is a kernel width parameter. We remove the edges whose weights are smaller than $0.7$. Then, we keep a certain proportion\textemdash say $q$\textemdash of edges in $\mathcal{G}_0$ unchanged to form the consensus graph. Finally, we add $(1-q)|\mathcal{E}_0|$ edges randomly to the consensus graph  to  form heterogeneous graphs $\mathcal{G}_1,\dots., \mathcal{G}_I$, where $|\mathcal{E}_0|$ is the number of edges in $\mathcal{G}_0$. We denote $1-q$ as the degree of heterogeneity.

\textit{2) Signal generation: }For the $i$-th local graph $\mathcal{G}_i$,  we generate $N_i$ smooth  signals from the distribution  \cite{dong2016learning}
\begin{shrinkfix}
    \begin{align}
        \mathbf{X}_{i,n} \sim \mathcal{N}\left(\mathbf{0}, (\mathbf{L}_i^{*})^{\dag} + \sigma_w^2\mathbf{I}\right),\;\;n =1,\dots,N_i,
    \label{signal-genegration}
    \end{align}
\end{shrinkfix}
where $\sigma_w$ is noise scale and $\mathbf{L}_i^{*}$ is the real Laplacian matrix of $\mathcal{G}_i$, and $\mathbf{X}_{i,n}$ is the $n$-th data (column) of $\mathbf{X}_i$.

\textit{3) Evaluation metric: } Four metrics are employed to evaluate the learned graphs, as listed in \eqref{MCC}. The first three metrics are used to evaluate classification tasks since determining whether two vertices are connected can be regarded as a binary classification problem. Specifically, $\mathrm{TP}$ is the true positive rate, $\mathrm{TN}$ is the true negative rate, $\mathrm{FP}$ is the false positive rate and $\mathrm{FN}$ denotes false negative rate. The third metric, F1-score ($\mathrm{FS}$) is the harmonic mean of $\mathrm{Precision}$ and $\mathrm{Recall}$. The last metric is the relative error ($\mathrm{RE}$), where $ \widehat{\mathbf{w}}$ is the learned graph, and $\mathbf{w}^*$ is the groundtruth. The results of all metrics for local graphs are the average of $I$ clients.
\begin{shrinkfix}
    \begin{align}
   & \mathrm{Precision} = \frac{\mathrm{TP}}{\mathrm{TP} + \mathrm{FP}},\;\;\;\;\mathrm{Recall} = \frac{\mathrm{TP}}{\mathrm{TP} +\mathrm{ FN}}, \notag\\
       & \mathrm{FS} = \frac{2\mathrm{TP}}{2\mathrm{TP} + \mathrm{FN} + \mathrm{FP}},\;\;\;      \mathrm{RE} = \frac{\lVert \widehat{\mathbf{w}} - \mathbf{w}^*\rVert_{2}}{\lVert \mathbf{w}^*\rVert_{2}}.
    \label{MCC}
    \end{align}
\end{shrinkfix}

\textit{4) Baselines: }We employ three categories of baselines. (\romannumeral1) We learn all local graphs independently (IGL), which enjoys full privacy as local clients do not release any data, including model updates. (\romannumeral2) We use two traditional FL algorithm, FedAvg \cite{mcmahan2017communication} and FedProx \cite{li2020federated1}, to solve the following problem
\begin{shrinkfix}
\begin{align}
\underset{\mathbf{w} \in \mathcal{W} }{\min} & \frac{1}{N_{all}}\sum_{i=1}^I {  \mathbf{z}_{i}^{\top}\mathbf{w} - \alpha\mathbf{1}^{\top} \log\left(\mathbf{S}\mathbf{w} + \zeta \mathbf{1}\right) + 2\beta\lVert \mathbf{w}\rVert_2^2}.
    \label{sec:ex-eq-1}
\end{align}
\end{shrinkfix}
The model learns a global graph collaboratively using data from all clients regardless of data heterogeneity.  (\romannumeral3) We leverage two personalized FL algorithms, MRMTL \cite{liu2022privacy} and Ditto \cite{li2021ditto}, to jointly learn all local graphs by further considering data heterogeneity. See supplementary materials for more details on how they are applied to multiple graph learning. For our algorithm, we use PPGL-L and PPGL-C to represent the learned local and consensus graphs, respectively.

\textit{5) Determination of parameters: } For our model, $\alpha$ is fixed as $1$, and $\beta$ is selected as the value achieving the best $\mathrm{FS}$ in IGL. Parameters $\nu$ and $\lambda$ are determined by grid search in $[1,100]$ and $[0.01, 1]$. On algorithmic side, we set $\xi = 0.9, K_i^{(t)} =K =  5, T = 50$. The stepsize $\eta_w$ is $0.005$, which is small enough to satisfy $\eta_w \leq 1/ L_{max}$. Moreover, we let $I = 5$ and  $\sigma_w = 0.1$ for synthetic data.  All parameters of baselines are selected as those achieving the best $\mathrm{FS}$ values.

\subsection{Synthetic Data}

\textit{1) Data size: }We first test the effect of data size. We fix $q$ as 0.5 and assume that all clients have the same data size, i.e., $N_i = N, i = 1,\dots, I$. We vary $N$ from 20 to 100, and the results are shown in Table.\ref{table-datasize}. It is observed that FedAvg and FedProx tend to obtain high $\mathrm{Recall}$ but low  $\mathrm{Precision}$, meaning that the corresponding graphs contain more edges. As $N$ increases, FedAvg yields the worst $\mathrm{FS}$, which is a comprehensive metric for evaluating the learned graph topology. The reason is that FedAvg learns a global graph using data from all clients. However, local graphs are heterogeneous in our experiment. Therefore, the learned global graph is far from local graphs. In sharp contrast, IGL learns local graphs independently and achieves better performance than FedAvg when $N$ is large. Our model also learns a personalized graph for each local client. Unlike IGL, we learn all local graphs jointly via a consensus graph, leading to better performance than IGL. Our model is also superior to Ditto and MRMTL\textemdash two PFL algorithms\textemdash since it is tailored for graph learning problems. Additionally, our model learns a consensus graph that captures the common structures of local graphs.

\begin{figure}[t] 
    \centering
       \subfloat[]{
       \includegraphics[width=0.5\linewidth]{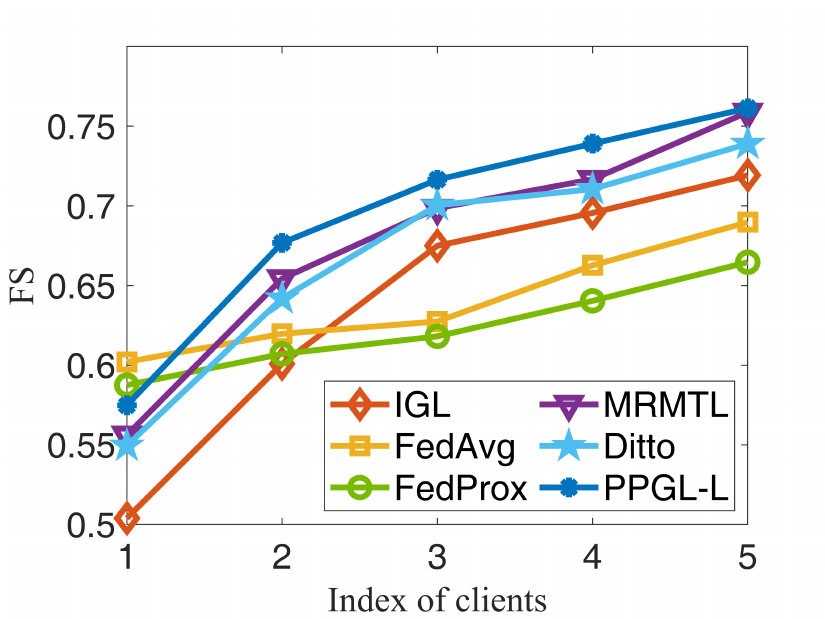}}
       \subfloat[]{
       \includegraphics[width=0.5\linewidth]{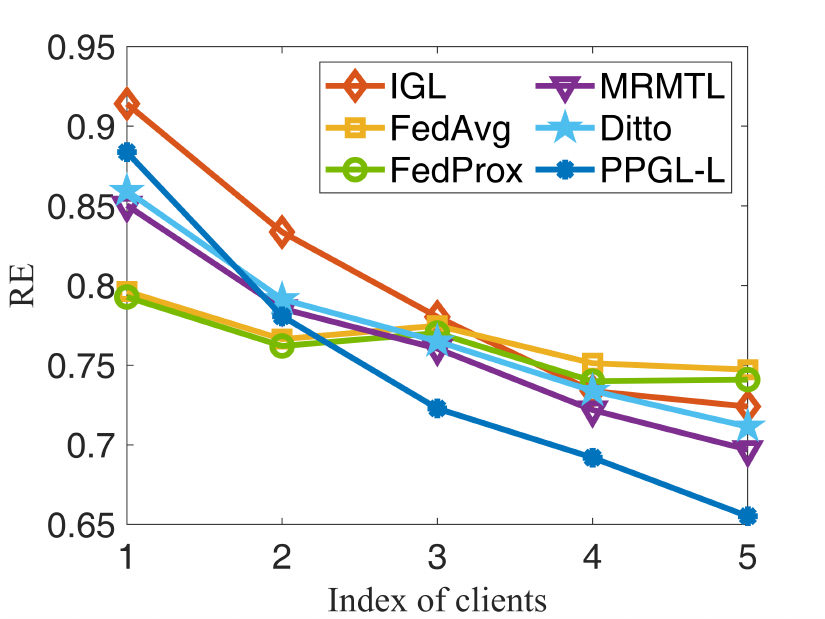}}
    	\caption{The results of varying data sizes of different clients. The data sizes  $N_1,\dots, N_5 $ are  20, 40, 60, 80, and 100, respectively.}
    	\label{Fig-datasize}
\end{figure}

We also test our model with different data sizes for different clients. We set $N_i$ to  20, 40, 60, 80, and 100 for $i=1,\dots,5$, respectively.  As illustrated in Fig.\ref{Fig-datasize}, clients with larger $N_i$ perform better than those with small $N_i$. However,  our model can improve the performance of clients with small $N_i$ because they can ``borrow" information from the clients with large $N_i$.

\begin{figure*}[t] 
    \centering
	  \subfloat[Ground truth]{
       \includegraphics[width=0.135\linewidth]{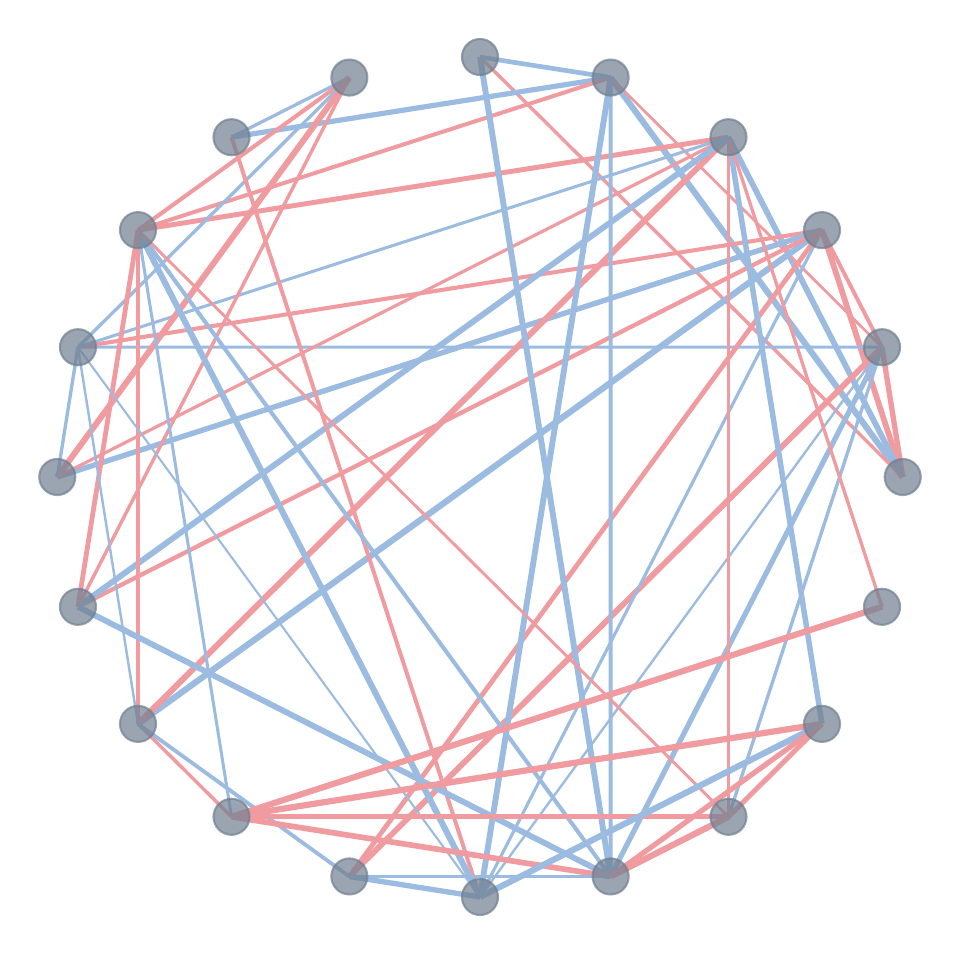}}
       \subfloat[PPGL]{
       \includegraphics[width=0.135\linewidth]{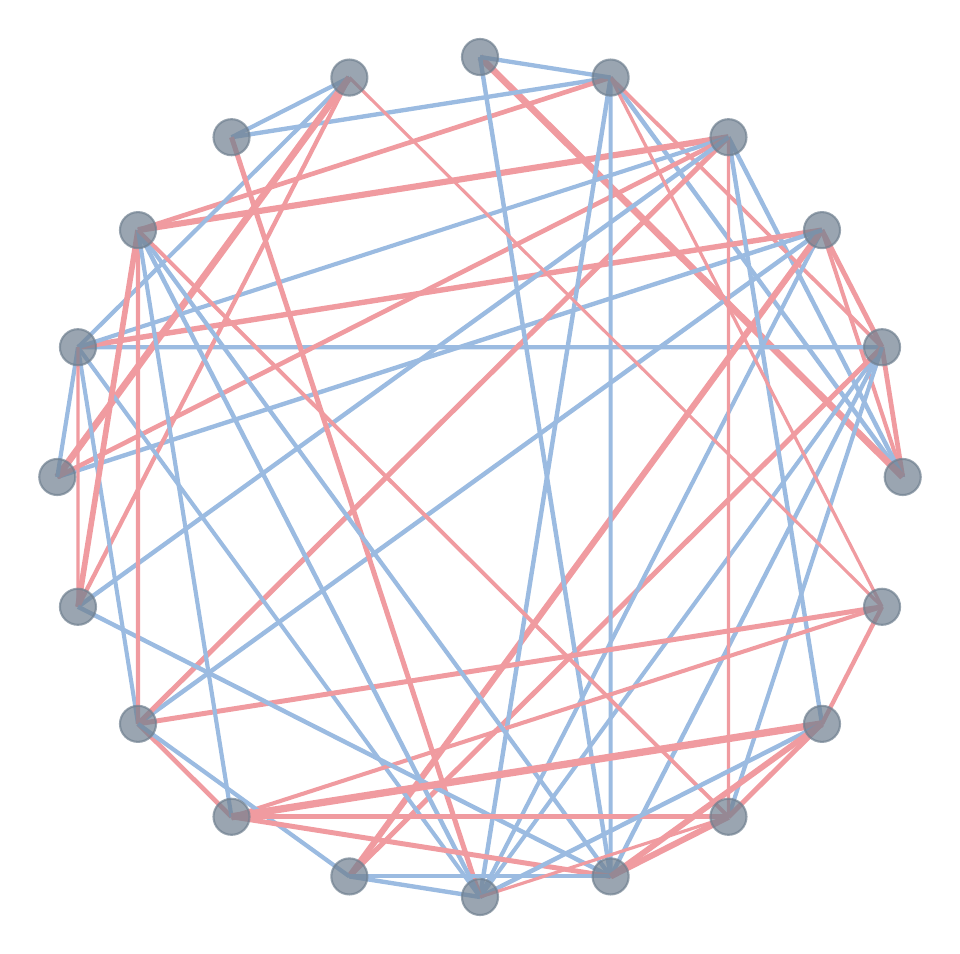}}
       \subfloat[IGL]{
       \includegraphics[width=0.135\linewidth]{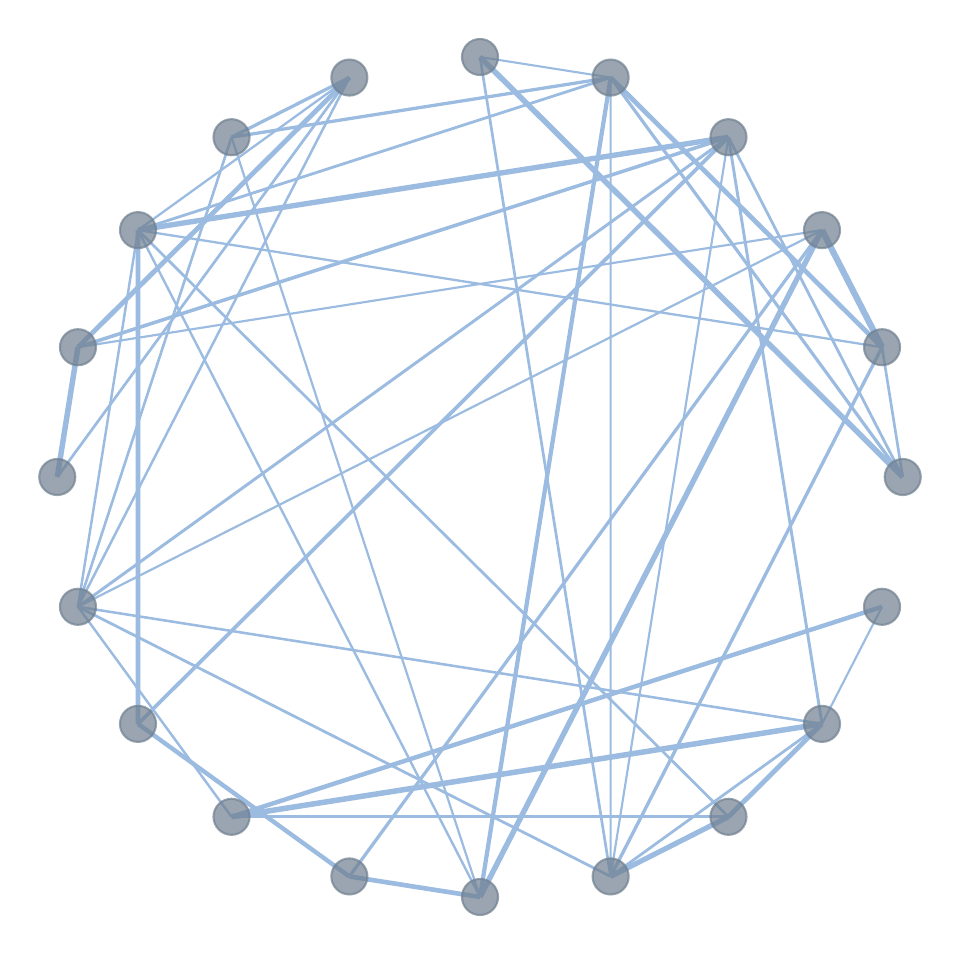}}
    \subfloat[FedAvg]{\includegraphics[width=0.135\linewidth]{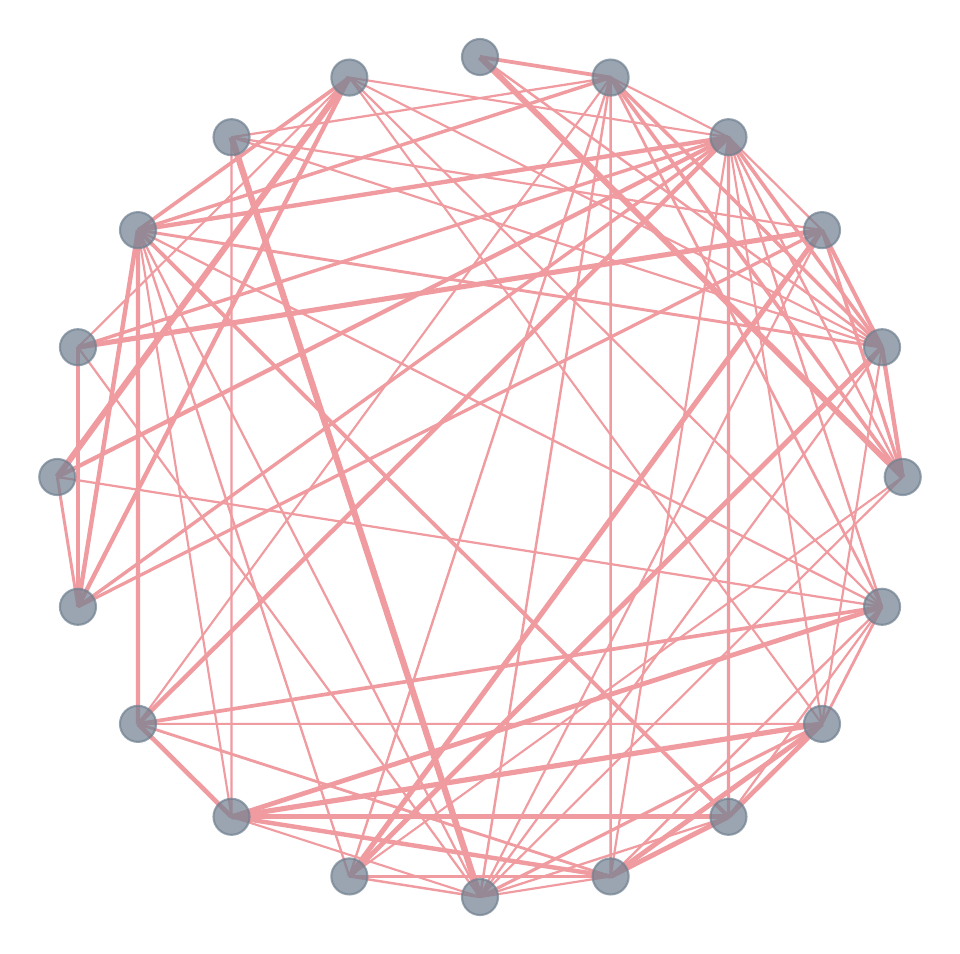}}
        \subfloat[FedProx]{
       \includegraphics[width=0.135\linewidth]{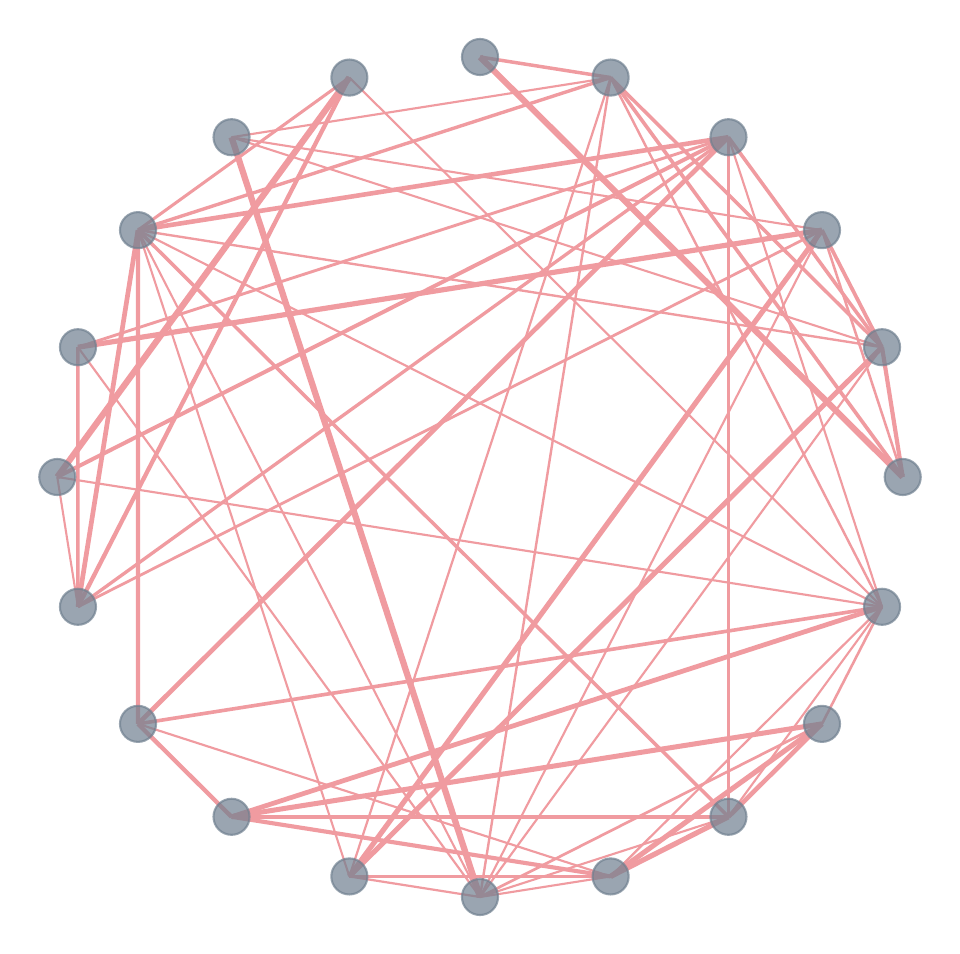}}
    \subfloat[Ditto]{
       \includegraphics[width=0.135\linewidth]{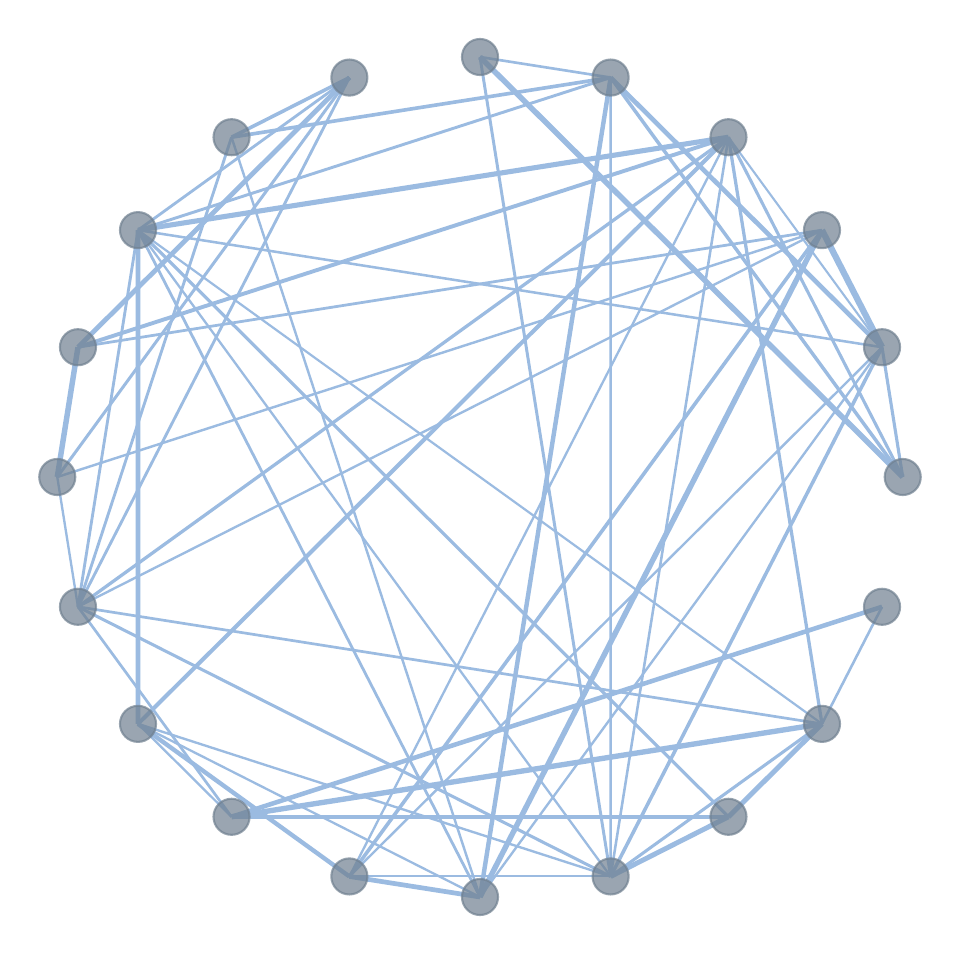}}
    \subfloat[MRMTL]{
       \includegraphics[width=0.135\linewidth]{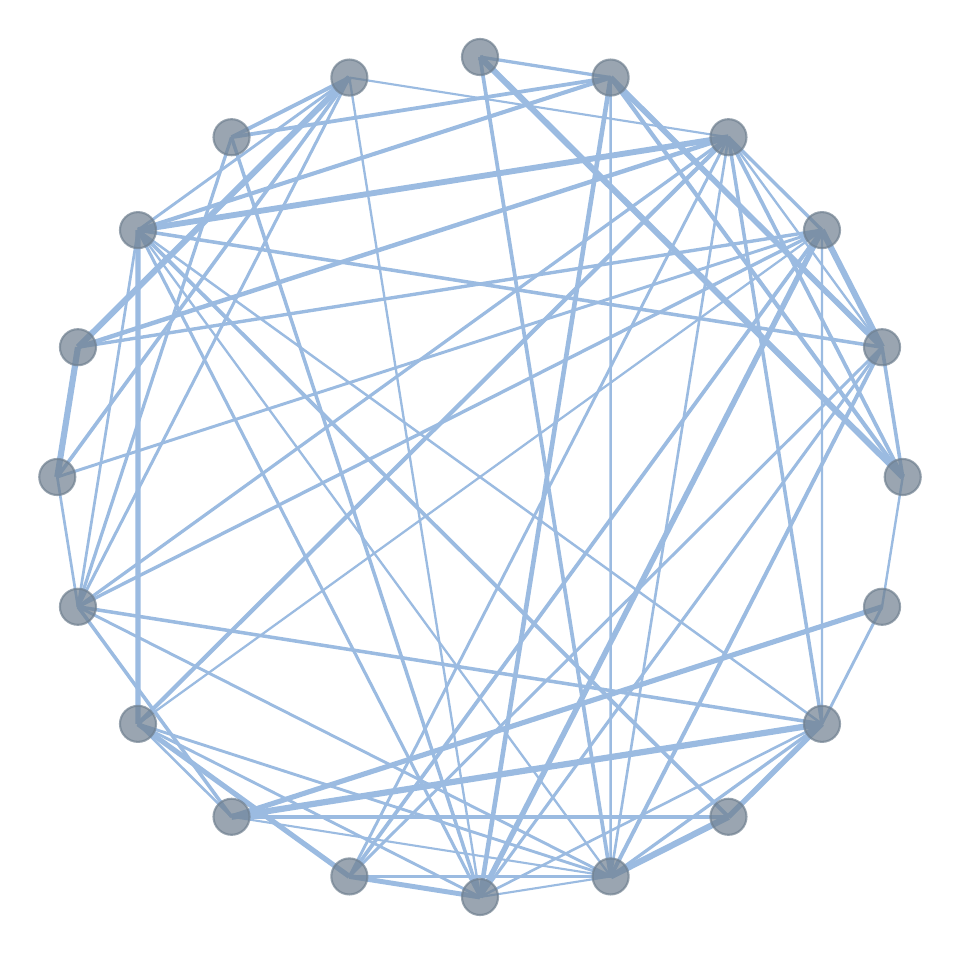}}

    	\caption{The  graphs of $\mathcal{C}_1$ learned by different methods when $N_i = 100, q = 0.5$. In (a)-(b), red edges are those in the consensus (global) graph, while blue edges are only in the local graphs. 
    	}
    	\label{Fig-visual}
\end{figure*}

Figure \ref{Fig-visual} provides the visualization of the learned graphs. We can observe that, compared with other methods, our model can effectively capture the common and specific structures of the real graph.

\textit{2) Data heterogeneity: }We next explore the effect of data heterogeneity. In this experiment, we fix $N_i = 50$ for all clients and vary $q$ from 0.3 to 0.9. As listed in Table \ref{table-heter},  when data heterogeneity is small (large $q$), FedAvg and FedProx achieve satisfactory performance since the consensus graph is close to local graphs. In this case, FedAvg and FedProx benefit from using all data to learn a global graph. However, with the increase in data heterogeneity, the personalized algorithms outperform FedAvg and FedProx. Moreover, our model outperforms Ditto and MRMTL due to the more reasonable consensus graph learning formulation. Our model obtains the best performance for all data heterogeneity since we learn personalized graphs and a consensus graph simultaneously.

\textit{3) The impact of adaptive weight schema: }Then, we test the effectiveness of the method of adjusting weights. We conduct an ablation experiment where the adaptive weight schema is removed from our model and let all local clients share the same weight. The local graphs and consensus graph output by this model are denoted as  ``PPFL-L-w/o" and ``PPFL-C-w/o", respectively. Two cases are considered, i.e., varying data sizes and varying data heterogeneity. For Case 1, we let $q$ be 0.5 and set $N_i$ to 20, 40, 60, 80, and 100 for $\mathcal{C}_1-\mathcal{C}_5$, respectively. For Case 2, we fix $N_i = 100$ and first generate a graph $\mathcal{G}_1$ for $\mathcal{C}_1$. The graphs of  $\mathcal{C}_2$-$\mathcal{C}_5$ are generated based on  $\mathcal{G}_1$ with $q$ equal to 0.8, 0.6, 0.4, and 0.2, respectively. The results are listed in Table \ref{ablation}. Note that removing adaptive weight schema from our model will degrade performance, indicating the effectiveness of the proposed auto-weighted model.

\begin{figure}[t] 
    \centering
       \includegraphics[width=0.7\linewidth]{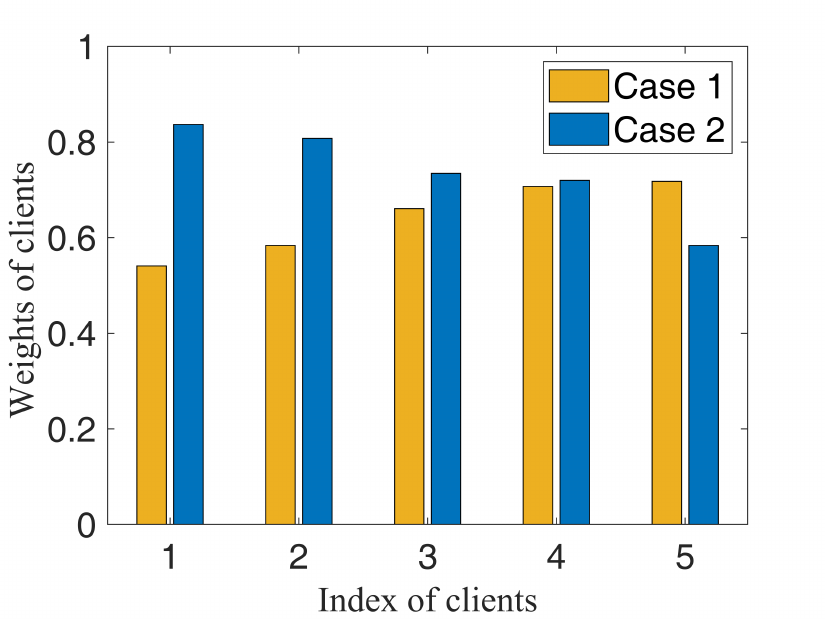}
    	\caption{The learned $\bm{\gamma}$ of different clients  
    	}
    	\label{Fig-client-weight}
\end{figure}

Moreover, the learned average weights of different clients are displayed in Fig.\ref{Fig-client-weight}, and two trends are observed. First, local graphs with large data sizes contribute more to the consensus graph.   Second, local graphs close to the consensus graph obtain larger weights. Thus, the weight adjustment strategy can effectively allocate a reasonable weight for each client.

\begin{table}[t]
\setlength{\tabcolsep}{1mm}
    \centering
    \caption{The results of removing adaptive weight schema. }
\small
    \begin{tabular}{ccccccc}
        \toprule
        & Methods & $\mathrm{Precision}\uparrow$  &$\mathrm{Recall}\uparrow$  &$\mathrm{FS}\uparrow$  & $\mathrm{RE}\downarrow$  \\
        \midrule
       \multirow{4}{*}{Case 1} & PPFL-L  &    0.691    &  \textbf{0.773} & 0.691  &0.705\\ 
        & PPFL-C  &     \textbf{0.743}   & 0.702 &  \textbf{0.722} & \textbf{0.689} \\ 
        & PPFL-L-w/o  &  0.618       & 0.751 & 0.678  &0.711 \\ 
        & PPFL-C-w/o  &   0.702      & 0.692 & 0.697  &0.701\\ 
        \hline
        
\multirow{4}{*}{Case 2}& PPFL-L  & 0.718 &  \textbf{0.875}  &0.789 & \textbf{0.605}  \\ 
& PPFL-C   &  \textbf{0.888} & 0.744  & \textbf{0.810} &0.611  \\ 
 & PPFL-L-w/o  & 0.704 & 0.857  &0.773 &0.625  \\ 
& PPFL-C-w/o   & 0.880 & 0.730  &0.798 &0.632  \\ 
        \bottomrule
    \end{tabular}
 \label{ablation}
\end{table}

% \setlength{\tabcolsep}{1mm}
% {
% 	\begin{table}[t]
% 		\centering
% 		\caption{The learned $\bm{\gamma}$ of different clients}
%   \small
% 		\begin{tabular}{cccccc}
% 			\toprule
% 			Index of clients &  1 & 2  &3   &4   &5\\
% 			\midrule
% 			Case 1           & 0.541 & 0.584  &0.661 &0.707 &0.718 \\ 
% 			Case 2 & 0.837 & 0.808  &0.720 &0.735 &0.584 \\ 
% 			\bottomrule
% 		\end{tabular}
% 	 \label{table-weight}
% 	\end{table}
% }

\begin{figure}[t] 
    \centering
	  \subfloat[Local graphs]{
       \includegraphics[width=0.5\linewidth]{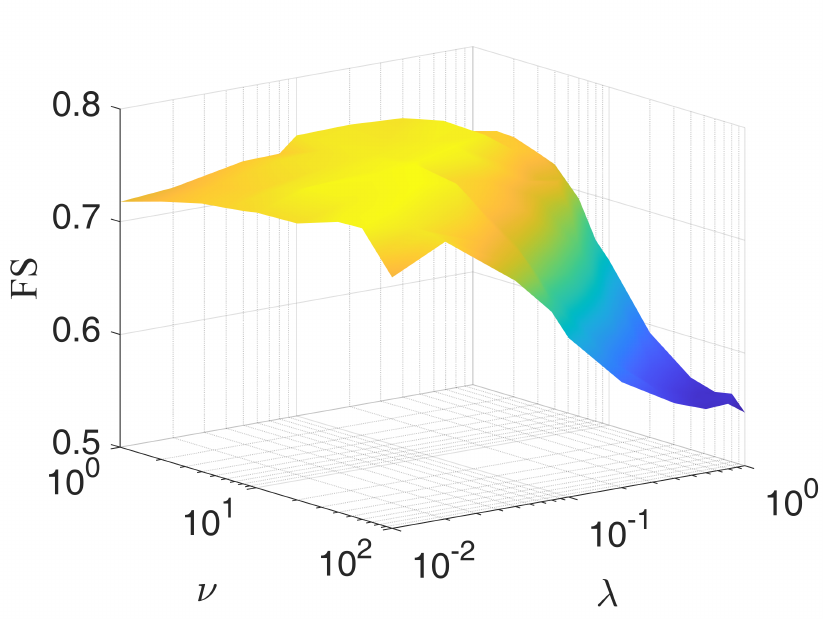}}
       \subfloat[The consensus graph]{
       \includegraphics[width=0.5\linewidth]{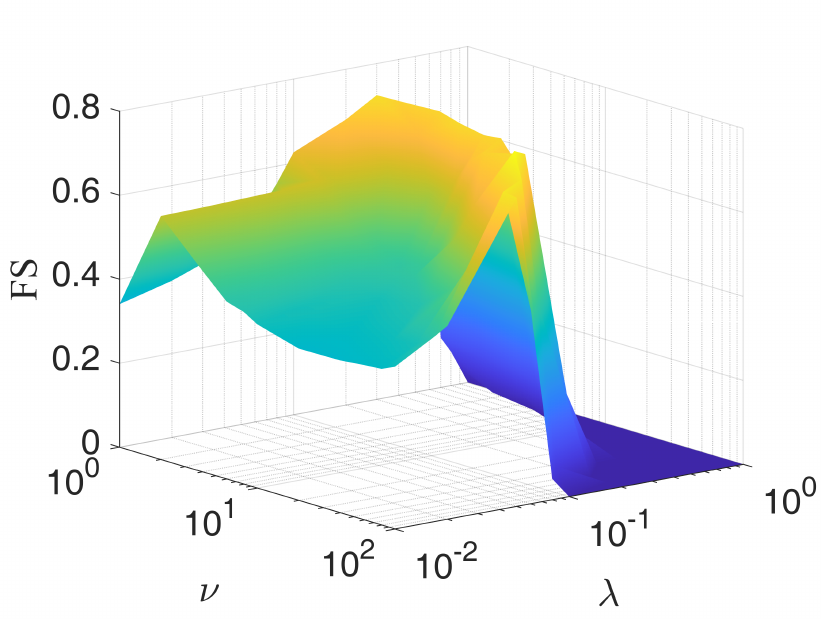}}
    	\caption{The results of parameter sensitivity.  
    	}
    	\label{Fig-parameter}
        \vspace{-1em}
\end{figure}

\textit{4) Parameter sensitivity: }In this experiment, we set $N_i = 100$ and $q = 0.5$, and evaluate the learning performance of different $\lambda$ and $\nu$. As displayed in Fig.\ref{Fig-parameter}, local graphs are less sensitive to the parameters than the consensus graph. When $\nu$ and $\lambda$ are too large, the performance of the consensus graph degrades rapidly.  However, there exist combinations of $\nu$ and $\lambda$ that achieve the highest $\mathrm{FS}$ for both local graphs and the consensus graph. We need to grid search for the optimal combination of $\nu$ and $\lambda$, especially for consensus graphs.

\begin{figure}[t] 
    \centering
	  \subfloat[Local graphs]{
       \includegraphics[width=0.5\linewidth]{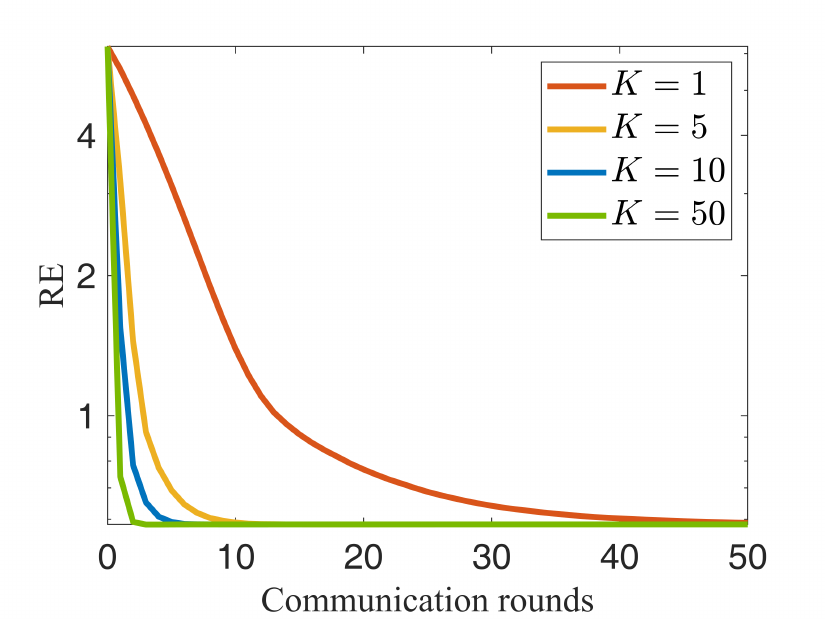}}
       \subfloat[The consensus graph]{
       \includegraphics[width=0.5\linewidth]{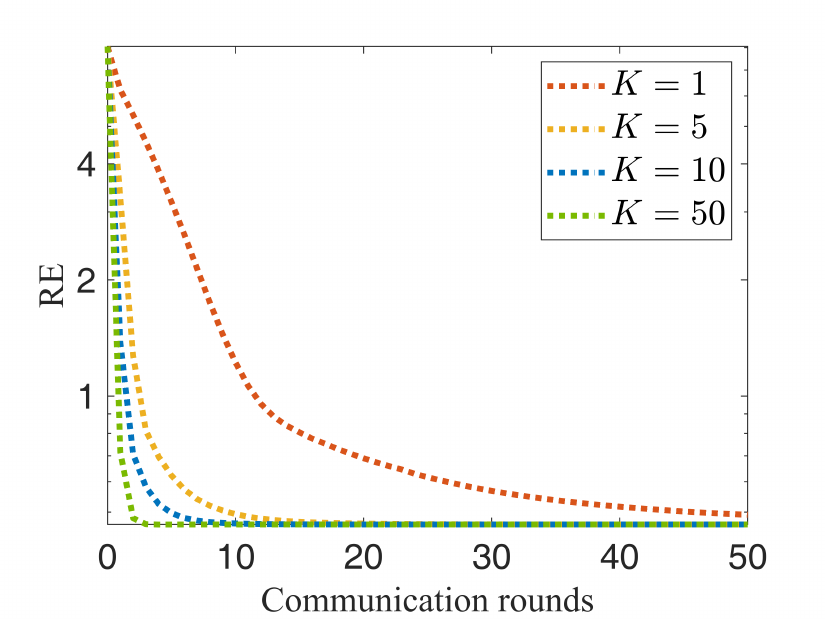}}
    	\caption{The convergence of the proposed algorithm.}
    	\label{Fig-convergence}
                \vspace{-1em}

\end{figure}

\textit{5) Algorithm convergence: }
\label{sec:exper-convergence}
Finally, we test the convergence of the proposed algorithms. Fig.\ref{Fig-convergence} shows that our algorithms with different $K$ can converge as stated in Theorem \ref{theorem-convergence}. When $K$ is small, fewer local updates are performed. In this case, more communication rounds between local clients and the central server are required. On the other hand, when $K$ is large, our algorithm only takes a few rounds to converge.

\subsection{Real-world Data}

\begin{figure*}[t] 
    \centering
	  \subfloat[IGL- $\mathcal{C}_2$ ]{
       \includegraphics[width=0.25\linewidth]{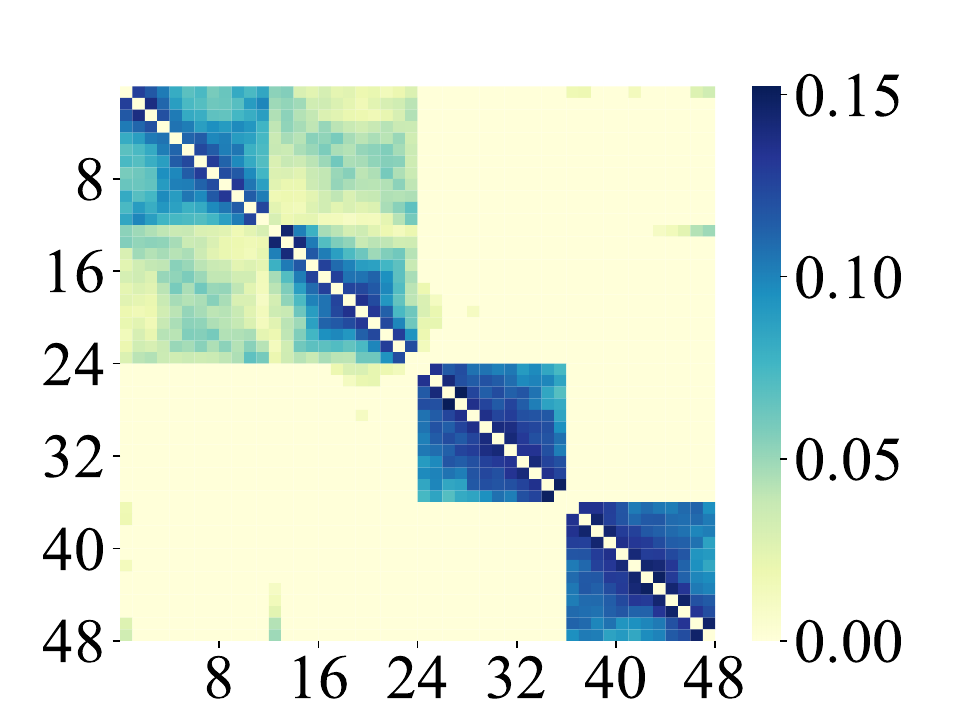}}
	  \subfloat[FedAvg]{       \includegraphics[width=0.25\linewidth]{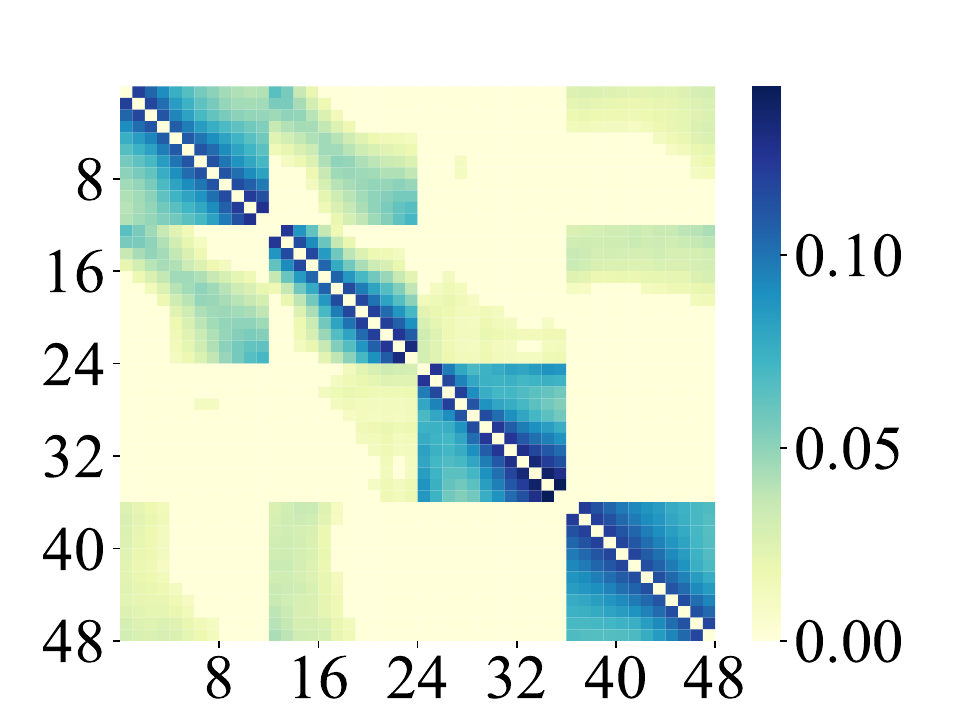}}
         \subfloat[PPGL-L- $\mathcal{C}_2$ ]{\includegraphics[width=0.25\linewidth]{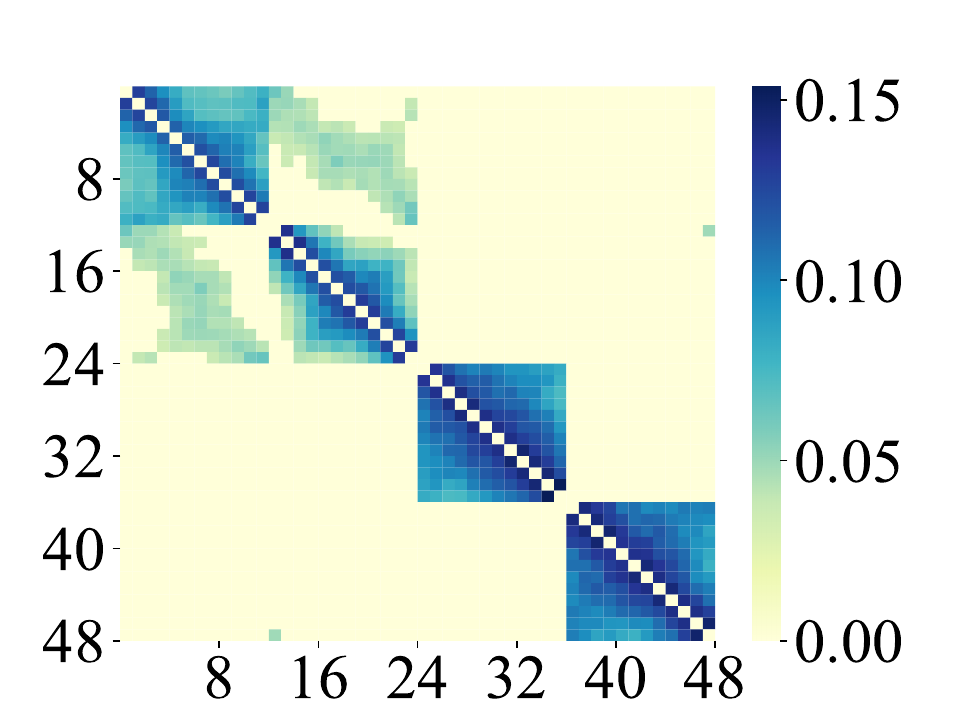}}
      \subfloat[PPGL-C]{\includegraphics[width=0.25\linewidth]{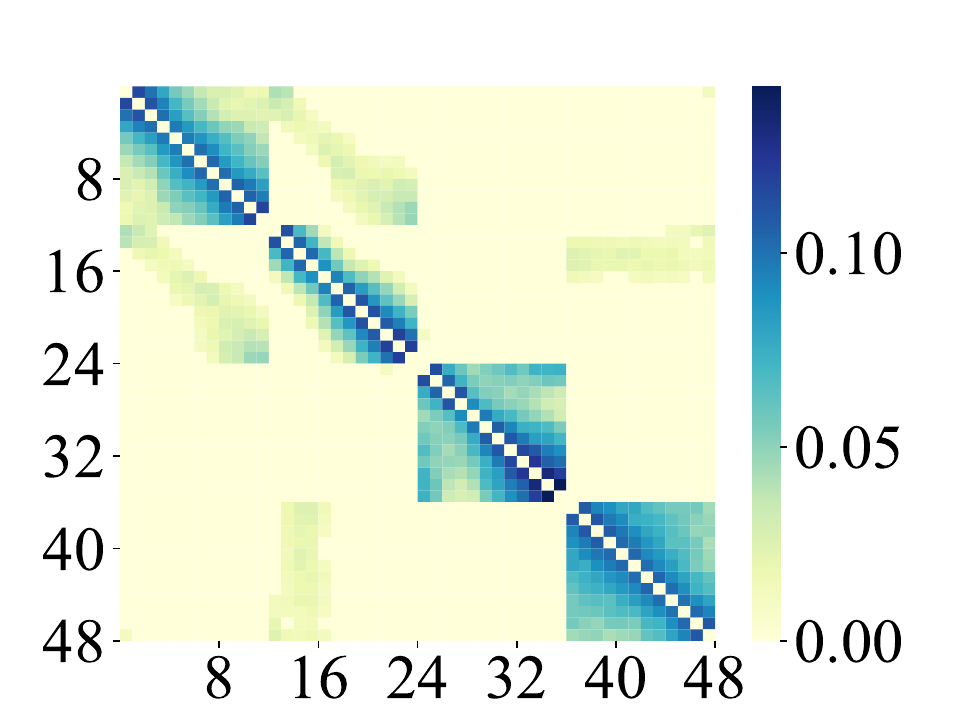}}

    	\caption{The figure depicts the communities in the learned graphs. The suffix $-\mathcal{C}_2$ stands for the local graphs of $\mathcal{C}_2$.}
 	\label{Fig-coil} 
\end{figure*}
\textit{1) COIL-20 data: }We employ the COIL-20 dataset\footnote{ https://www.cs.columbia.edu/CAVE/software/softlib/coil-20.php}, which is a collection of gray-scale images including 20 objects taken from 360 degrees, to learn the relationships between these images. Each object has 72 images (five degrees an image) of size $32\times32$. We randomly select four objects and divide the images of each object into three views. Each view contains images taken in consecutive 60 degrees, e.g., $[0^{\circ}, 55^{\circ}]$. Therefore, each view contains 48 images taken from 4 objects, i.e., 12 images per object. A basic assumption is that the 48 images of different views are defined on the same node set. The image itself is taken as graph signals, i.e.,  $\mathbf{X}_i\in \mathbb{R}^{48\times 1024}$.  Data distributions of different views are heterogeneous because they are from different shooting angles. Furthermore, we assume that photos from three views are taken by three photographers, and they are reluctant to share their photos, which is the concern of this study. Our goal is to learn a graph representing the relationships among these 48 images for each view, i.e., $I=3$, under data silos. The three graphs should share some common structures because they come from the same objects. Moreover, there should be four communities in the relation graphs since all images belong to four objects. Therefore, to evaluate the learned graphs, we employ the  Louvain algorithm \cite{fortunato2010community} to detect communities in the learned graphs. Three metrics are leveraged to evaluate the detection results, i.e., normalized mutual information ($\mathrm{NMI}$),  Fowlkes and Mallows index ($\mathrm{FMI}$) \cite{han2011data}, and Rand Index ($\mathrm{RI}$).  The labels of images are taken as the ground truth, and the results of local graphs are the average of three views. As displayed in Table \ref{table-coil},  the Louvain algorithm successfully finds all communities in the consensus graph learned by our models. Instead, some mistakes are made in the graphs learned by IGL and FedAvg. This is expected since the consensus graph captures the common structure of local graphs, which may remove noisy edges of different views.  Furthermore, the local graphs learned by our model also outperform those of other models since our proposed model is tailored for graph learning problems. As shown in   Fig.\ref{Fig-coil}, the communities can be clearly observed in the consensus graph. However, more confusing edges appear in the graphs of IGL and FedAvg.

\setlength{\tabcolsep}{0.7mm}
{
	\begin{table}[t]
		\centering
		\caption{The community detection results of graphs learned by different methods.}
  \small
		\begin{tabular}{cccccccccccc}
			\toprule
			&  IGL   & FedAvg & FedProx& MRMTL &Ditto & PPGL-L & PPGL-C   \\
			\midrule
			$\mathrm{NMI}\uparrow$  & 0.779  &0.745  &0.767  &0.795  &0.786 & {0.865}& \textbf{1}   \\ 
			$\mathrm{RI}\uparrow$  & 0.855 &0.843    &0.851  &0.891  &0.883& {0.919} &\textbf{1}   \\ 
			$\mathrm{FMI}\uparrow$  & 0.755  &0.736  &0.745  &0.773 &0.768& {0.852}  &\textbf{1}\\ 			
			\bottomrule
		\end{tabular}
	 \label{table-coil}
	\end{table}
}

\begin{table}[t]
		\centering
		\caption{Comparison between the graphs learned from the autism and non-autism groups.}
  \small
		\begin{tabular}{ccc}
			\toprule
			 &   \#Edges & Average Edge weights  \\
			\midrule
               Autism   & 114 & 0.091  \\
			Non-autism  & 125 & 0.109  \\ 
			\bottomrule
		\end{tabular}
	 \label{table-comparison-autism}
\end{table}

 \begin{figure}[t] 
    \centering
	  \subfloat[Sagittal]{
       \includegraphics[width=0.35\linewidth]{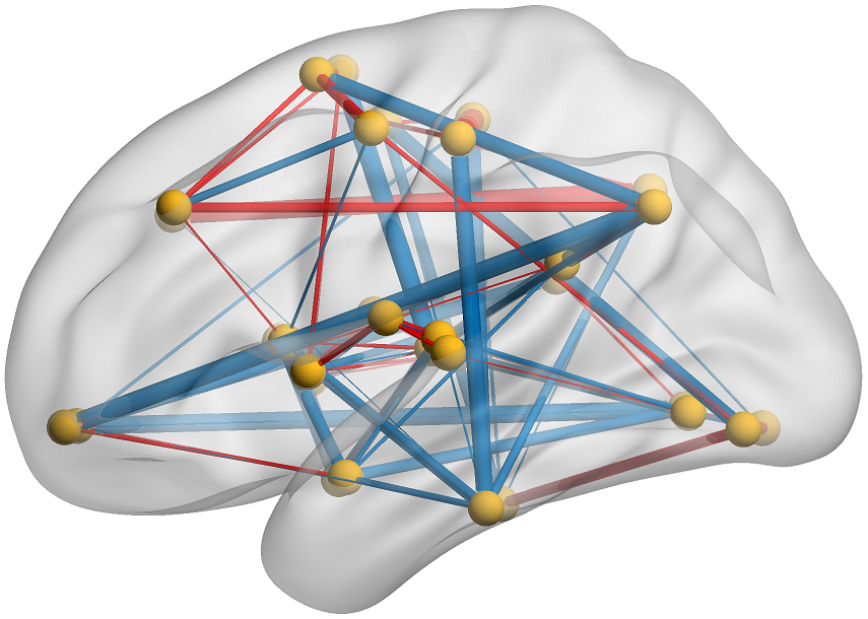}}
	  \subfloat[Axial]{
       \includegraphics[width=0.25\linewidth]{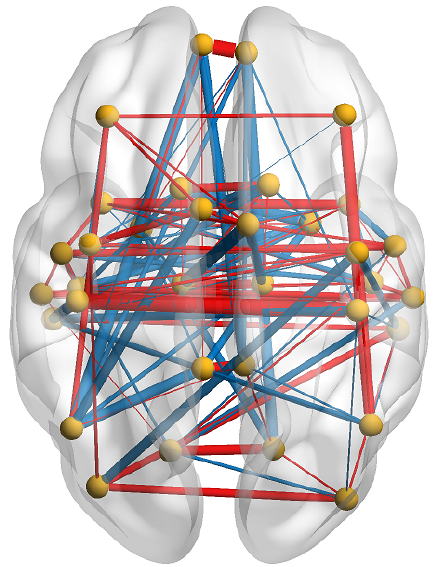}}
       \subfloat[Coronal]{
       \includegraphics[width=0.28\linewidth]{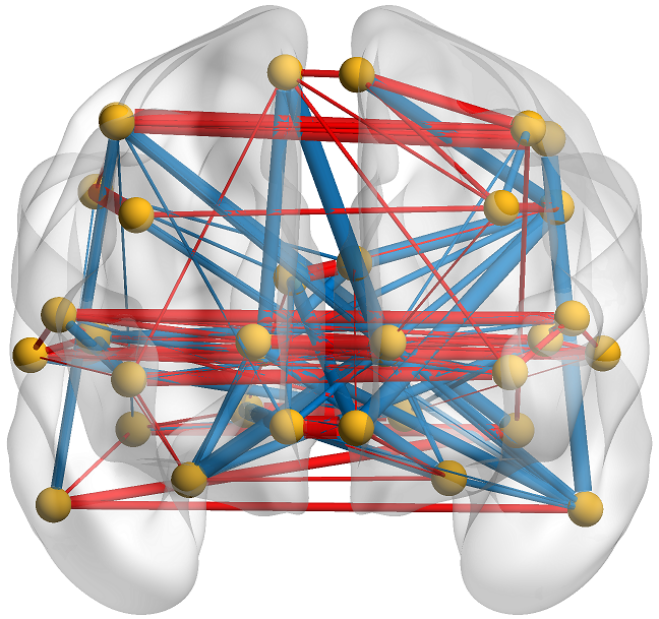}}

       \subfloat[Sagittal]{
       \includegraphics[width=0.35\linewidth]{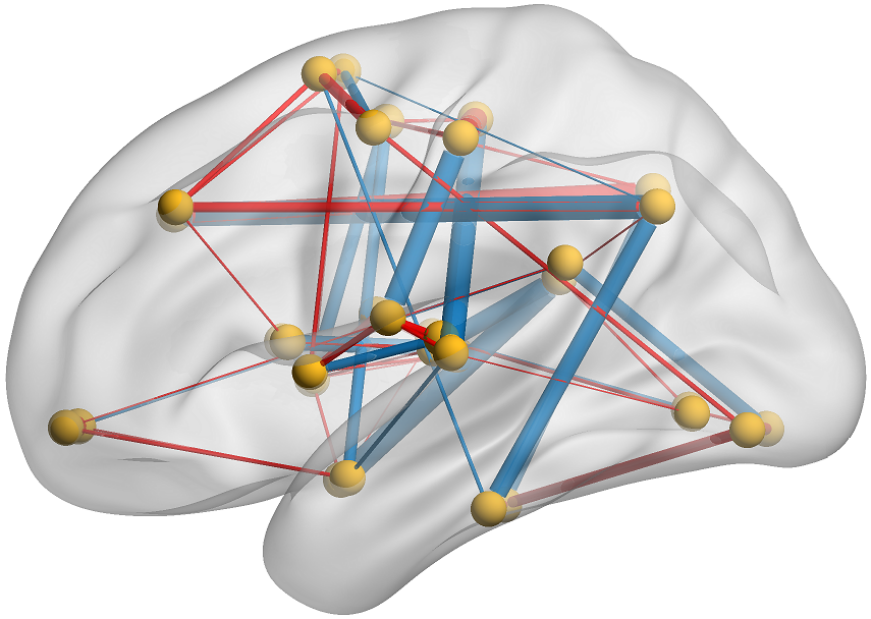}}
       \subfloat[Axial]{
       \includegraphics[width=0.25\linewidth]{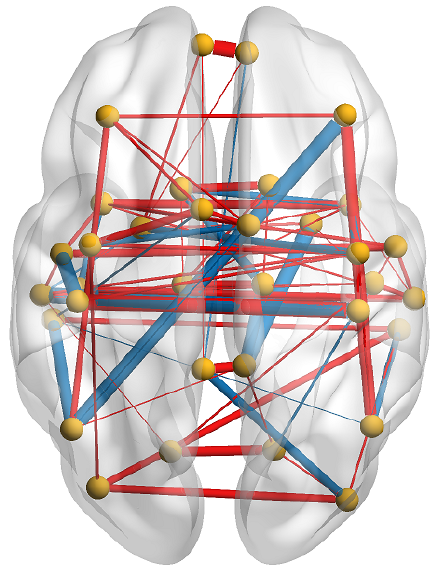}}
       \subfloat[Coronal]{
       \includegraphics[width=0.28\linewidth]{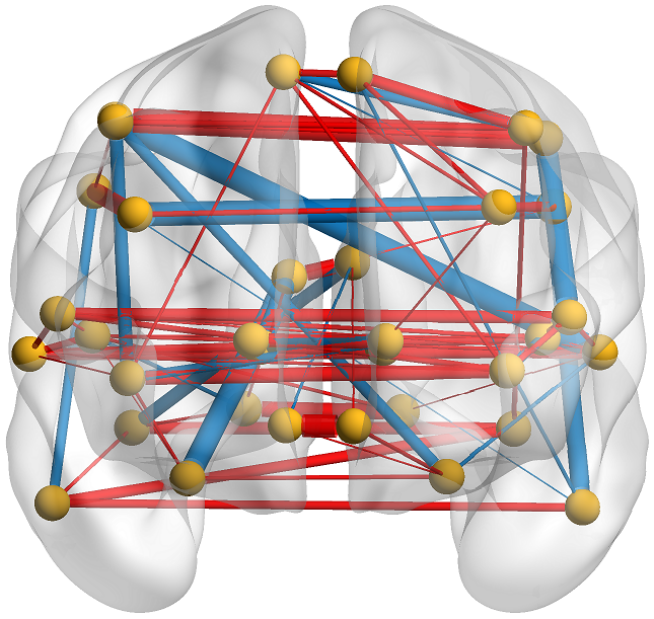}}
    	\caption{ The figure depicts our learned graphs from three views. The top and bottom rows show the graphs of non-autistic and autistic subjects, respectively.  The red edges are those in the consensus graph, while the blue edges are those unique edges in local graphs.}
    	\label{Fig-brain}
\end{figure}

\textit{2) Medical data: } We finally employ blood-oxygenation-level-dependent (BOLD) time series extracted from fMRI data to learn brain functional connectivity graphs. The basic assumption is that autism may affect brain functional connectivity. If we can learn about the differences in brain functional connectivity graphs between autistic and non-autistic people, it may help us better understand the mechanism of autism. However, medical data are privacy-sensitive, and their transmission to an unreliable central server is usually prohibited. In this experiment, the used dataset\footnote{http://preprocessed-connectomes-project.org/abide/} contains 539  autistic individuals and 573 typical controls, from which we select fifty autistic and fifty non-autistic subjects. Each client (subject) contains 176 signals (the length of the BOLD time series is 176). Following \cite{pu2021learning}, we select 34 functional regions of interest from 90 standard regions of the Anatomical Automatic Labeling (AAL) template, indicating that $\mathbf{X}_i \in\mathbb{R}^{34\times 176}, i = 1,\dots,100$. Then, we use our method to learn the connectivity graphs of the 34 regions, and the edges with weights less than 0.01 are removed. We should mention that the purpose of this experiment is to show that our method can learn graphs that reflect the impact of autism on brain functional connectivity. Next, we will show that the selected 34 regions are sufficient to achieve this goal.

Table \ref{table-comparison-autism} displays the average number of edges and their average weights of the graphs learned from the autism and non-autism subjects. It is observed the average number of the graphs from autism subjects is smaller than those of non-autism subjects. Furthermore, the graphs learned from autism subjects exhibit smaller edge weights than those of non-autism subjects, indicating weaker connectivity between different functional regions. This is consistent with the study \cite{kana2011disrupted}, which shows that autism may cause underconnectivity of brain functional regions. Furthermore, Fig.\ref{Fig-brain} depicts the learned graphs using our model. It is observed that graphs of autism and non-autism individuals share many common edges learned in the consensus graph. On the other hand, there are some noticeable topological changes between the graphs of autism and non-autism individuals. These edges reflect changes in the functional connectivity caused by autism, which may aid in diagnosing autism.  In the future, a domain expert may help interpret the results from a medical perspective.

\begin{comment}
\end{comment}

\section{Conclusion}
\label{sec:Conclusion}
In this paper,  we proposed a framework to learn graphs under data silos. In our framework, we jointly learned a personalized graph for each client to handle data heterogeneity, and a consensus graph for all clients to capture global information. Then, we devised a tailored algorithm in which all private data are processed in local clients to preserve data privacy. Convergence analyses for the proposed algorithm were provided.  Extensive experiments showed that our approach can efficiently learn graphs in the target scenario. Future research may include generalizing our framework to other graph learning models except for the smoothness assumption.

\appendices
\section{Proof of Theorem \ref{theo-estimation-error}}
\label{appendix-0}
We first provide the following lemmas necessary to prove the Theorem \ref{theo-estimation-error}.

\begin{lemma}
   The  function $h(\mathbf{w} )$ is $4\beta$-strongly convex.
    \label{lemma-1-for-theorem-1}
\end{lemma}
\begin{proof}
    For any $\mathbf{w}\geq 0$, the Hessian matrix of $h(\mathbf{w})$ w.r.t. $\mathbf{w}$ is $ \mathbf{H}= 4\beta \mathbf{I}+ \alpha \mathbf{S}^{\top}\mathrm{diag}\left((\mathbf{S}\mathbf{w})^{.(-2)}\right)\mathbf{S}$. It is not difficult to check that $\mathbf{H} \succ 4\beta\mathbf{I}$. Thus, $h(\mathbf{w})$ is $4\beta$-strongly convex.
    \label{proof-lemma-1-for-theorem-1}
\end{proof}

\begin{lemma}
For any $\mathbf{Y} = \left[\mathbf{y}_1,\dots, \mathbf{y}_{I+1} \right]\in\mathbb{R}^{p\times(I+1)}$, we have $R(\mathbf{Y})\leq \left(\nu \sqrt{I\omega_{\max}(\mathbf{L}_m)} + \sqrt{p}\right) \lVert \mathbf{Y}\rVert_{\mathrm{F}}$.
    \label{lemma-2-for-theorem-1}
\end{lemma}
\begin{proof}
Let $\widetilde{\mathbf{y}}_j\in\mathbb{R}^{I+1} $ be the $j$-th row vector of the matrix $\mathbf{Y}$, it is not difficult to obtain
\begin{shrinkfix}
\begin{align}
    &R(\mathbf{Y}) = \nu\sum_{i=1}^I \lVert \mathbf{y}_i - \mathbf{y}_{I+1} \rVert_{2} + \lVert \mathbf{y}_{I+1} \rVert_{1} \notag\\
    \leq &\nu \sqrt{I} \sqrt{\sum_{i=1}^I \lVert \mathbf{y}_i - \mathbf{y}_{I+1} \rVert_{2}^2} + \sqrt{p}\lVert \mathbf{y}_{I+1}  \rVert_{2} \notag\\
     = & \nu \sqrt{I} \sqrt{\sum_{j=1}^p\widetilde{\mathbf{y}}_j\mathbf{L}_m \widetilde{\mathbf{y}}_j^{\top}} + \sqrt{p} \lVert \mathbf{y}_{I+1}  \rVert_{2}\notag\\
     \leq &\nu \sqrt{I}\sqrt{ \sum_{j=1}^p \omega_{\max}(\mathbf{L}_m)  \lVert\widetilde{\mathbf{y}}_j\rVert_2^2} + \sqrt{p}\lVert \mathbf{Y} \rVert_{\mathrm{F}}\notag\\
     = &\left( \nu \sqrt{I\omega_{\max}(\mathbf{L}_m)}+ \sqrt{p}\right)  \lVert \mathbf{Y}\rVert_{\mathrm{F}},
\end{align}
\end{shrinkfix}
where the first inequality holds due to the basic inequality $\frac{\sum_{i=1}^n a_i}{n} \leq \sqrt{\frac{\sum_{i=1}^n a_i^2}{n}}$ for $a_i\geq 0$ and norm inequality $\lVert \mathbf{a}\rVert_1 \leq \sqrt{p}\lVert \mathbf{a}\rVert_2$ for $\mathbf{a}\in\mathbb{R}^p$. The second inequality holds due to the definition of the $\ell_2$ matrix norm and $\lVert \mathbf{y}_{I+1} \rVert_{2} \leq \lVert \mathbf{Y}\rVert_{\mathrm{F}}$. Finally, we complete the proof.
    \label{proof-lemma-2-for-theorem-1}
\end{proof}

With the two lemmas, we start the proof of Theorem \ref{theo-estimation-error}. First, the objective function \eqref{sec:con-eq-1} can be rewritten in a matrix form. Let $\mathbf{Z} = [\mathbf{z}_1,\dots,\mathbf{z}_I, \mathbf{0}]\in \mathbb{R}^{p\times (I+1)} $, and we have
\begin{align}
\underset{\mathbf{W}\geq 0 }{\mathrm{min}}\,\, (1/N)\left\langle \mathbf{Z}, \mathbf{W}\right\rangle + H(\mathbf{W}) + \lambda R(\mathbf{W}),
\label{eq1-appdx0}
\end{align}
where $H(\mathbf{W}) = \sum_{i=1}^I h(\mathbf{w}_i)$. Since $\widehat{\mathbf{W}}$ is the solution of \eqref{eq1-appdx0}, we have 
\begin{shrinkfix}
\begin{align}
    &(1/N)\left\langle \mathbf{Z}, \widehat{\mathbf{W}}\right\rangle + H(\widehat{\mathbf{W}}) +  \lambda R(\widehat{\mathbf{W}})\notag\\ 
    \leq &(1/N)\left\langle \mathbf{Z}, \mathbf{W}^*\right\rangle + H(\mathbf{W}^*) + \lambda R(\mathbf{W}^*)
     \label{append-equ-2}.
\end{align}
\end{shrinkfix}
Accordingly, we yield that 
\begin{shrinkfix}
\begin{align}
    &(1/N)\left\langle \mathbf{Z}^*, \widehat{\mathbf{W}} - \mathbf{W}^*\right\rangle + H(\widehat{\mathbf{W}}) - H(\mathbf{W}^*)  \notag\\
    \leq  & \lambda R(\mathbf{W}^*) -  \lambda R(\widehat{\mathbf{W}}) + (1/N) \left\langle \mathbf{E}, \mathbf{W}^* - \widehat{\mathbf{W}} \right\rangle
     \label{append-equ-3}, 
\end{align}
\end{shrinkfix}
where $\mathbf{E} = \left[\mathbf{e}_1,\dots,\mathbf{e}_I, \mathbf{0}\right]\in \mathbb{R}^{p\times (I+1)}$ is the error matrix, and $\mathbf{Z}^* = [\mathbf{z}_1^*,\dots,\mathbf{z}_I^*, \mathbf{0}]\in \mathbb{R}^{p\times (I+1)}$ is the real pairwise distance matrix.

Then, we define a variable $v = \sum_{i=1}^I\sum_{j = 1}^p \frac{1}{\sigma_e^2} (\mathbf{e}_{i}[j])^2 = \frac{1}{\sigma_e^2}\lVert \mathbf{E} \rVert_{\mathrm{F}}^2$ which follows a chi-squared distribution with the degree of freedom as $pI$. Using the Wallace inequality \cite{wallace1959bounds}, for  $\delta > 0$, we have
\begin{shrinkfix}
\begin{align}
    \mathrm{Pr}(v\geq pI+ \delta) \leq \exp\left( -\frac{1}{2}\left(\delta - pI\log\left(1 + \frac{\delta}{pI} \right)\right)\right),
    \label{append-equ-4}
\end{align}
\end{shrinkfix}
which will lead to
\begin{shrinkfix}
\begin{align}
&\mathrm{Pr}\left((1/N)\lVert \mathbf{E}\rVert_{\mathrm{F}}\leq ({\sigma_e}/{N})\sqrt{pI+ \delta}\right)\notag\\
\geq &1 -\exp\left( -\frac{1}{2}\left(\delta - pI\log\left(1 + \frac{\delta}{pI} \right)\right)\right).
\label{append-equ-5}
\end{align}
\end{shrinkfix}
According to \eqref{append-equ-5}, with probability at least $1 -\exp\left( -\frac{1}{2}\left(\delta - pI\log\left(1 + \frac{\delta}{pI} \right)\right)\right)$, we obtain
\begin{shrinkfix}
\begin{align}
    &(1/N)\left\langle \mathbf{E}, \mathbf{W}^* - \widehat{\mathbf{W}} \right\rangle\leq (1/N) \lVert \mathbf{E}\rVert_{\mathrm{F}} \lVert \mathbf{W}^* - \widehat{\mathbf{W}}\rVert_{\mathrm{F}}\notag\\
    &\leq({\sigma_e}/{N})\sqrt{pI+ \delta} \lVert \mathbf{W}^* - \widehat{\mathbf{W}}\rVert_{\mathrm{F}}
    \label{append-equ-6}.
\end{align}
\end{shrinkfix}

Next, we focus on $H$ related terms in \eqref{append-equ-3} and yield 
\begin{shrinkfix}
\begin{align} 
     &H(\widehat{\mathbf{W}}) -H (\mathbf{W}^*)= \sum_{i=1}^I h(\widehat{\mathbf{w}}_i) - h(\mathbf{w}_i^*)\notag\\
    \geq &\sum_{i=1}^I \left\langle \nabla h(\mathbf{w}_i^*),  \widehat{\mathbf{w}}_i - \mathbf{w}_i^*  \right\rangle+ 2\beta \left\lVert \widehat{\mathbf{w}}_i - \mathbf{w}_i^* \right\rVert_2^2\notag\\
    \geq & \sum_{i=1}^I  -\left \lVert \nabla h(\mathbf{w}_i^*) \right\rVert_2 \left\lVert  \widehat{\mathbf{w}}_i - \mathbf{w}_i^*\right\rVert_2 + 2\beta \left\lVert \widehat{\mathbf{w}}_i - \mathbf{w}_i^* \right\rVert_2^2 \notag\\
    \geq&  -C_h \sqrt{I} \left\lVert  \widehat{\mathbf{W}} - \mathbf{W}^*\right\rVert_{\mathrm{F}} + 2\beta \left\lVert  \widehat{\mathbf{W}} - \mathbf{W}^*\right\rVert_{\mathrm{F}}^2.
    \label{append-equ-9}
\end{align}
\end{shrinkfix}
The first inequality holds due to that $h$ is a $4\beta$- strongly convex function, as stated in Lemma \ref{lemma-1-for-theorem-1}. The second inequality holds due to Cauchy-Schwarz inequality. The last inequality holds due to
Assumption \ref{assumption-1-1} and the fact that $\sum_{i=1}^I \lVert \widehat{\mathbf{w}}_i - \mathbf{w}_i^* \rVert_{2} \leq \sqrt{I}\lVert\widehat{\mathbf{W}} - \mathbf{W}^* \rVert_{\mathrm{F}}$.

For $\mathbf{Z}^*$ related terms, we have 
\begin{shrinkfix}
\begin{align} 
&(1/N)  \left\langle \mathbf{Z}^*,  \widehat{\mathbf{W}} - \mathbf{W}^*\right\rangle 
\geq -(1/N)   \left\lVert\mathbf{Z}^*\right\rVert_{\mathrm{F}}\left\lVert  \widehat{\mathbf{W}} - \mathbf{W}^*\right\rVert_{\mathrm{F}} \notag\\
&\geq -\left(\sqrt{I} C_z/N \right) \left\lVert  \widehat{\mathbf{W}} - \mathbf{W}^*\right\rVert_{\mathrm{F}}.
    \label{append-equ-10}
\end{align}
\end{shrinkfix}
The second inequality holds due to Assumption \ref{assumption-0}.

Finally, we focus on the $R$ related terms and have
\begin{shrinkfix}
\begin{align} 
&R(\mathbf{W}^*) - R(\widehat{\mathbf{W}})\notag\\
= &\nu\sum_{i=1}^I \left(\lVert \mathbf{w}_i^* - \mathbf{w}^*_{\mathrm{con}}\rVert_2  - \lVert \widehat{\mathbf{w}}_i - \widehat{\mathbf{w}}_{\mathrm{con}}\rVert_2 \right)\notag\\
&+ \lVert \mathbf{w}^*_{\mathrm{con}} \rVert_1 - \lVert \widehat{\mathbf{w}}_{\mathrm{con}} \rVert_1 \notag\\
\leq &\nu\sum_{i=1}^I \lVert (\mathbf{w}_i^* -\widehat{\mathbf{w}}_i) - (\mathbf{w}^*_{\mathrm{con}} -\widehat{\mathbf{w}}_{\mathrm{con}})\lVert_2 +  \lVert \mathbf{w}_{\mathrm{con}}^*-\widehat{\mathbf{w}}_{\mathrm{con}} \rVert_1 \notag\\
=&R(\mathbf{W}^* - \widehat{\mathbf{W}})
\leq \left( \nu \sqrt{I\omega_{\max}(\mathbf{L}_m)}+ \sqrt{p}\right)  \lVert\mathbf{W}^* -\widehat{\mathbf{W}}\rVert_{\mathrm{F}},
    \label{append-equ-11}
\end{align}
\end{shrinkfix}
%where the first inequality holds $\lVert \mathbf{a}_1 \rVert_q - \lVert \mathbf{a}_2 \rVert_q \leq \lVert \mathbf{a}_1 - \mathbf{a}_2\rVert_q$ for $q=1,2$. 
where the last inequality holds due to Lemma \ref{lemma-2-for-theorem-1}.

Plugging \eqref{append-equ-6}, \eqref{append-equ-9},  \eqref{append-equ-10}, and \eqref{append-equ-11} into \eqref{append-equ-3}, it is not difficult to yield 
\begin{shrinkfix}
\begin{align} 
 &2\beta \left\lVert  \widehat{\mathbf{W}} - \mathbf{W}^*\right\rVert_{\mathrm{F}}^2 
 \leq ({\sigma_e}/{N})\sqrt{pI+ \delta} \left\lVert \mathbf{W}^* - \widehat{\mathbf{W}}\right\rVert_{\mathrm{F}}\notag\\
 & + C_h \sqrt{I} \left\lVert  \widehat{\mathbf{W}} - \mathbf{W}^*\right\rVert_{\mathrm{F}} + \left(\sqrt{I} C_z/N \right) \left\lVert  \widehat{\mathbf{W}} - \mathbf{W}^*\right\rVert_{\mathrm{F}}\notag\\
 &+  \lambda\left( \nu \sqrt{I\omega_{\max}(\mathbf{L}_m)}+ \sqrt{p}\right)  \left\lVert\widehat{\mathbf{W}} - \mathbf{W}^*\right\rVert_{\mathrm{F}}.
    \label{append-equ-12}
\end{align}
\end{shrinkfix}
Finally, combining \eqref{append-equ-12} and \eqref{eq-lambda-selection}, we can obtain the conclusion of Theorem \ref{theo-estimation-error}.

\section{Proof of Proposition \ref{fact-1}}
\label{appendix-1}
The proof is mainly from \cite{saboksayr2021online} but with some modifications. For two vectors $\mathbf{y}$ and $\mathbf{y}^{\prime}$ in $\widetilde{\mathcal{W}}$, we have 
\begin{shrinkfix}
\begin{align}
&\left\lVert\nabla_{\mathbf{y}} f^{(t)}_i(\mathbf{y},\mathbf{w}_{\mathrm{con}}^{(t)}) - \nabla_{\mathbf{y}^{\prime}} f^{(t)}_i(\mathbf{y}^{\prime},\mathbf{w}_{\mathrm{con}}^{(t)})\right\rVert_2 \notag\\
= &\left\lVert \left(4\beta + \rho\gamma_i^{(t)}\right)  (\mathbf{y} - \mathbf{y}^{\prime}) - \alpha \mathbf{S}^{\top}\left(\frac{1}{\mathbf{S\mathbf{y}} + \zeta\mathbf{1}} - \frac{1}{\mathbf{S\mathbf{y}^{\prime}} + \zeta\mathbf{1}}\right) \right\rVert_2\notag\\
\leq& \left(4\beta + \rho\gamma_i^{(t)}\right)\left\lVert\mathbf{y} - \mathbf{y}^{\prime}\right\rVert_2 + \alpha \left\lVert  \mathbf{S}^{\top}\right\rVert_2 \left\lVert\frac{1}{\mathbf{S\mathbf{y}}+ \zeta\mathbf{1}} - \frac{1}{\mathbf{S\mathbf{y}^{\prime}}+ \zeta\mathbf{1}}\right\rVert_2 \notag\\
\leq &\left(4\beta+\rho\gamma_i^{(t)}\right) \left\lVert\mathbf{y} - \mathbf{y}^{\prime}\right\rVert_2 + \frac{\alpha\lVert \mathbf{S}\rVert_2}{\zeta^2}\left\lVert\mathbf{S}\mathbf{y} -  \mathbf{S}\mathbf{y}^{\prime} \right\rVert_2 \notag\\
\leq& \left(4\beta+\rho\gamma_i^{(t)}\right)\left\lVert\mathbf{y} - \mathbf{y}^{\prime}\right\rVert_2 + \frac{\alpha \lVert \mathbf{S}\rVert_2^2}{\zeta^2} \left\lVert\mathbf{y} -  \mathbf{y}^{\prime}\right\rVert_2 \notag\\
 = & \left(4\beta + \frac{2\alpha(d-1)}{\zeta^2} + \rho\gamma_i^{(t)}\right)\lVert \mathbf{y} -\mathbf{y}^{\prime} \rVert_2\notag\\
 :=& L_i^{(t)} \lVert \mathbf{y} -\mathbf{y}^{\prime} \rVert_2.
\label{sec:con-eq-3}
\end{align}
\end{shrinkfix}
The first inequality holds due to the triangle inequality, while the second equality holds due to  Lemma 1 in \cite{saboksayr2021online}. From \eqref{sec:con-eq-3}, we can conclude that $f_i^{(t)}(\mathbf{w}_i,\mathbf{w}_{\mathrm{con}}^{(t)})$ is  $L_i^{(t)}$-Lipschitz smooth w.r.t. $\mathbf{w}_i$ on $\widetilde{\mathcal{W}}$. On the other hand, the strong convexity of $ f_i^{(t)} (\mathbf{w}_i, \mathbf{w}_{\mathrm{con}}^{(t)})$ can be  proved in the same way as Lemma \ref{lemma-1-for-theorem-1}.

\section{Proof of Theorem \ref{theorem-convergence}}
\label{appendix-2}
First, let us provide a lemma that is essential to the proof.
\begin{lemma}
(\cite{nie2010efficient})
For any two positive constants $a, b>0$, we have
\begin{align}
\sqrt{a} - \frac{a}{2\sqrt{b}} \leq \sqrt{b} - \frac{b}{2\sqrt{b}}.
    \label{eq-lemma-uv}
\end{align}
    \label{lemma-uv}
\end{lemma}

\begin{proof}
For $a,b>0$, it is not difficult to derive that 
\begin{align}
    &\left(\sqrt{a} -  \sqrt{b}\right)^2\geq 0\notag\\
     \implies& a - 2\sqrt{ab} + b\geq 0 \notag\\
    \implies& \sqrt{a} - \frac{a}{2\sqrt{b}} \leq \frac{\sqrt{b}}{2}\notag\\
     \implies &\sqrt{a} - \frac{a}{2\sqrt{b}} \leq \sqrt{b} - \frac{b}{2\sqrt{b}}.
    \label{eq-lemma-uv-1}
\end{align} 
The proof of Lemma \ref{lemma-uv} completes.
\end{proof}

Then, we start to prove Theorem \ref{theorem-convergence}. Let us consider the $t$-th communication round. According to Proposition \ref{fact-1}, 
the objective function $f_i^{(t)}(\mathbf{w}_i,\mathbf{w}_{\mathrm{con}}^{(t)})$ is $L_i^{(t)}$-Lipschitz smooth and  $S_i^{(t)}$-strongly convex w.r.t. $\mathbf{w}_{i}$. Based on Assumption \ref{assumption-2} and the stepsize $\eta_w$ defined in Assumption \ref{assumption-3}, the accelerated projected gradient descent algorithm \eqref{sec:alg-eq-2-0}-\eqref{sec:alg-eq-2-2} is proved to converge to the optimal minimum \cite{nesterov2013introductory}. Thus, after enough $K_i^{(t)}$ iterations, we can obtain $\mathbf{w}_{i}^{(t+1)}$ such that
\begin{align}
f_i^{(t)}(\mathbf{w}_{i}^{(t+1)}, \mathbf{w}_{\mathrm{con}}^{(t)}) \leq f_i^{(t)}(\mathbf{w}_{i}^{(t)}, \mathbf{w}_{\mathrm{con}}^{(t)}).
\label{eq-theo-converge-0}
\end{align}

On the other hand, when updating $\mathbf{w}_{\mathrm{con}}$ in the $t$-th communication round, the  $\mathbf{w}_{\mathrm{con}}^{(t+1)}$ output by proximal operator \eqref{sec:alg-eq-45} is the solution of problem \eqref{sec:alg-eq-3}, i.e., 
\begin{align}
\mathbf{w}_{\mathrm{con}}^{(t+1)} = \underset{\mathbf{w}_{\mathrm{con}}}{\mathrm{argmin}}\,\, \sum_{i=1}^{I} f_i^{(t)}(\mathbf{w}_i^{(t+1)}, \mathbf{w}_{\mathrm{con}})  + \lambda\lVert \mathbf{w}_{\mathrm{con}}\rVert_1.\notag
\end{align}
Therefore, we have 
\begin{align}
&\sum_{i=1}^{I} f_i^{(t)}(\mathbf{w}_i^{(t+1)}, \mathbf{w}_{\mathrm{con}}^{(t+1)})  + \lambda\lVert \mathbf{w}_{\mathrm{con}}^{(t+1)}\rVert_1\notag\\
\leq& \sum_{i=1}^{I} f_i^{(t)}(\mathbf{w}_i^{(t+1)}, \mathbf{w}_{\mathrm{con}}^{(t)})  + \lambda\lVert \mathbf{w}_{\mathrm{con}}^{(t)}\rVert_1. 
\label{eq-theo-converge-1}
\end{align}

Combining \eqref{eq-theo-converge-0} and \eqref{eq-theo-converge-1}, we can obtain 
\begin{align}
&\sum_{i=1}^{I} f_i^{(t)}(\mathbf{w}_i^{(t+1)}, \mathbf{w}_{\mathrm{con}}^{(t+1)})  + \lambda\lVert \mathbf{w}_{\mathrm{con}}^{(t+1)}\rVert_1\notag\\
\leq& \sum_{i=1}^{I} f_i^{(t)}(\mathbf{w}_i^{(t)}, \mathbf{w}_{\mathrm{con}}^{(t)})  + \lambda\lVert \mathbf{w}_{\mathrm{con}}^{(t)}\rVert_1.
\label{eq-theo-converge-3-0}
\end{align}
Bring the updated $\gamma_i^{(t)}$ in \eqref{sec:alg-eq-5-new} to \eqref{eq-theo-converge-3-0}, we obtain 
\begin{align}
& \sum_{i=1}^{I} g_i(\mathbf{w}_i^{(t+1)}) +  \rho \frac{\lVert\mathbf{w}_i^{(t+1)} -  \mathbf{w}_{\mathrm{con}}^{(t+1)}\rVert_2^2}{2\lVert\mathbf{w}_i^{(t)} -  \mathbf{w}_{\mathrm{con}}^{(t)}\rVert_2} +\lambda\lVert \mathbf{w}_{\mathrm{con}}^{(t+1)}\rVert_1 \notag\\
\leq &\sum_{i=1}^{I} g_i(\mathbf{w}_i^{(t)}) +  \rho\frac{\lVert\mathbf{w}_i^{(t)} -  \mathbf{w}_{\mathrm{con}}^{(t)}\rVert_2^2}{2\lVert\mathbf{w}_i^{(t)} -  \mathbf{w}_{\mathrm{con}}^{(t)}\rVert_2}+\lambda\lVert \mathbf{w}_{\mathrm{con}}^{(t)}\rVert_1.
\label{eq-theo-converge-2}
\end{align}
According to Lemma \ref{lemma-uv}, we have
\begin{align}
&\sum_{i=1}^{I}\lVert\mathbf{w}_i^{(t+1)} -  \mathbf{w}_{\mathrm{con}}^{(t+1)}\rVert_2 - \sum_{i=1}^{I}\frac{\lVert\mathbf{w}_i^{(t+1)} -  \mathbf{w}_{\mathrm{con}}^{(t+1)}\rVert_2^2}{2\lVert\mathbf{w}_i^{(t)} -  \mathbf{w}_{\mathrm{con}}^{(t)}\rVert_2} \notag\\
\leq &\sum_{i=1}^{I}\lVert\mathbf{w}_i^{(t)} -  \mathbf{w}_{\mathrm{con}}^{(t)}\rVert_2 - \sum_{i=1}^{I}\frac{\lVert\mathbf{w}_i^{(t)} -  \mathbf{w}_{\mathrm{con}}^{(t)}\rVert_2^2}{2\lVert\mathbf{w}_i^{(t)} -  \mathbf{w}_{\mathrm{con}}^{(t)}\rVert_2}.
\label{eq-theo-converge-3}
\end{align}
Summing \eqref{eq-theo-converge-2} and \eqref{eq-theo-converge-3}, we have 
\begin{align}
& \sum_{i=1}^{I} g_i(\mathbf{w}_i^{(t+1)}) +\rho\lVert\mathbf{w}_i^{(t+1)} -  \mathbf{w}_{\mathrm{con}}^{(t+1)}\rVert_2 +\lambda\lVert \mathbf{w}_{\mathrm{con}}^{(t+1)}\rVert_1  
\notag\\
\leq &\sum_{i=1}^{I} g_i(\mathbf{w}_i^{(t)}) +\rho\lVert\mathbf{w}_i^{(t)} -  \mathbf{w}_{\mathrm{con}}^{(t)}\rVert_2 +\lambda\lVert \mathbf{w}_{\mathrm{con}}^{(t)}\rVert_1.
\label{eq-theo-converge-4}
\end{align}
Thus, the update of the $t$-th communication round will monotonically decrease the objective function of \eqref{sec:con-eq-1}. As $t$ increases, the optimization will converge. When the convergence reaches, the KKT condition of problem \eqref{sec:con-eq-1} will hold, i.e., Algorithm \ref{alg:FGL}
at least converges to a local optimal minimum.

% \section{Proof of Theorem \ref{theo-pd-analysis}}
% \label{appendix-3}
% \input{Appendix/appendix3}

% \section{Proof of Corollary \ref{coro-1}}
% \label{appendix-4}
% \input{Appendix/appendix4}

\bibliographystyle{IEEEtran}
\bibliography{IEEEabrv, references}

\begin{thebibliography}{10}
\providecommand{\url}[1]{#1}
\csname url@samestyle\endcsname
\providecommand{\newblock}{\relax}
\providecommand{\bibinfo}[2]{#2}
\providecommand{\BIBentrySTDinterwordspacing}{\spaceskip=0pt\relax}
\providecommand{\BIBentryALTinterwordstretchfactor}{4}
\providecommand{\BIBentryALTinterwordspacing}{\spaceskip=\fontdimen2\font plus
\BIBentryALTinterwordstretchfactor\fontdimen3\font minus \fontdimen4\font\relax}
\providecommand{\BIBforeignlanguage}[2]{{%
\expandafter\ifx\csname l@#1\endcsname\relax
\typeout{** WARNING: IEEEtran.bst: No hyphenation pattern has been}%
\typeout{** loaded for the language `#1'. Using the pattern for}%
\typeout{** the default language instead.}%
\else
\language=\csname l@#1\endcsname
\fi
#2}}
\providecommand{\BIBdecl}{\relax}
\BIBdecl

\bibitem{dong2019learning}
X.~Dong, D.~Thanou, M.~Rabbat, and P.~Frossard, ``Learning graphs from data: A signal representation perspective,'' \emph{{IEEE} Signal Process. Mag.}, vol.~36, no.~3, pp. 44--63, 2019.

\bibitem{von2007tutorial}
U.~Von~Luxburg, ``A tutorial on spectral clustering,'' \emph{Stat. Comput.}, vol.~17, pp. 395--416, 2007.

\bibitem{wu2020comprehensive}
Z.~Wu, S.~Pan, F.~Chen, G.~Long, C.~Zhang, and S.~Y. Philip, ``A comprehensive survey on graph neural networks,'' \emph{{IEEE} Trans. Neural Netw. Learn. Syst.}, vol.~32, no.~1, pp. 4--24, 2020.

\bibitem{mateos2019connecting}
G.~Mateos, S.~Segarra, A.~G. Marques, and A.~Ribeiro, ``Connecting the dots: Identifying network structure via graph signal processing,'' \emph{{IEEE} Signal Process. Mag.}, vol.~36, no.~3, pp. 16--43, 2019.

\bibitem{friedman2008sparse}
J.~Friedman, T.~Hastie, and R.~Tibshirani, ``Sparse inverse covariance estimation with the graphical lasso,'' \emph{Biostatistics}, vol.~9, no.~3, pp. 432--441, 2008.

\bibitem{egilmez2017graph}
H.~E. Egilmez, E.~Pavez, and A.~Ortega, ``Graph learning from data under laplacian and structural constraints,'' \emph{{IEEE} J. Sel. Topics Signal Process.}, vol.~11, no.~6, pp. 825--841, 2017.

\bibitem{ying2020nonconvex}
J.~Ying, J.~V. de~Miranda~Cardoso, and D.~Palomar, ``Nonconvex sparse graph learning under laplacian constrained graphical model,'' \emph{Proc. Adv. Neural Inf. Process. Syst.}, vol.~33, pp. 7101--7113, 2020.

\bibitem{cai2024fast}
J.-F. Cai, J.~V. de~Miranda~Cardoso, D.~Palomar, and J.~Ying, ``Fast projected newton-like method for precision matrix estimation under total positivity,'' \emph{Proc. Adv. Neural Inf. Process. Syst.}, vol.~36, 2024.

\bibitem{ortega2018graph}
A.~Ortega, P.~Frossard, J.~Kova{\v{c}}evi{\'c}, J.~M. Moura, and P.~Vandergheynst, ``Graph signal processing: Overview, challenges, and applications,'' \emph{Proc. {IEEE}}, vol. 106, no.~5, pp. 808--828, 2018.

\bibitem{dong2016learning}
X.~Dong, D.~Thanou, P.~Frossard, and P.~Vandergheynst, ``Learning {Laplacian} matrix in smooth graph signal representations,'' \emph{{IEEE} Trans. Signal Process.}, vol.~64, no.~23, pp. 6160--6173, 2016.

\bibitem{kalofolias2016learn}
V.~Kalofolias, ``How to learn a graph from smooth signals,'' in \emph{Proc. Int. Conf. Artif. Intell. Stat., AISTATS}.\hskip 1em plus 0.5em minus 0.4em\relax PMLR, 2016, pp. 920--929.

\bibitem{pu2021learning}
X.~Pu, T.~Cao, X.~Zhang, X.~Dong, and S.~Chen, ``Learning to learn graph topologies,'' \emph{Proc. Adv. Neural Inf. Process. Syst.}, vol.~34, pp. 4249--4262, 2021.

\bibitem{kairouz2021advances}
P.~Kairouz, H.~B. McMahan, B.~Avent, A.~Bellet, M.~Bennis, A.~N. Bhagoji, K.~Bonawitz, Z.~Charles, G.~Cormode, R.~Cummings \emph{et~al.}, ``Advances and open problems in federated learning,'' \emph{Found. Trends Mach. Learn.}, vol.~14, no. 1--2, pp. 1--210, 2021.

\bibitem{rieke2020future}
N.~Rieke, J.~Hancox, W.~Li, F.~Milletari, H.~R. Roth, S.~Albarqouni, S.~Bakas, M.~N. Galtier, B.~A. Landman, K.~Maier-Hein \emph{et~al.}, ``The future of digital health with federated learning,'' \emph{NPJ Digit. Med}, vol.~3, no.~1, pp. 1--7, 2020.

\bibitem{li2020federated}
T.~Li, A.~K. Sahu, A.~Talwalkar, and V.~Smith, ``Federated learning: Challenges, methods, and future directions,'' \emph{{IEEE} Signal Process. Mag.}, vol.~37, no.~3, pp. 50--60, 2020.

\bibitem{yang2019survey}
T.~Yang, X.~Yi, J.~Wu, Y.~Yuan, D.~Wu, Z.~Meng, Y.~Hong, H.~Wang, Z.~Lin, and K.~H. Johansson, ``A survey of distributed optimization,'' \emph{Annu. Rev. Control}, vol.~47, pp. 278--305, 2019.

\bibitem{mcmahan2017communication}
B.~McMahan, E.~Moore, D.~Ramage, S.~Hampson, and B.~A. y~Arcas, ``Communication-efficient learning of deep networks from decentralized data,'' in \emph{Proc. Int. Conf. Artif. Intell. Stat., AISTATS}.\hskip 1em plus 0.5em minus 0.4em\relax PMLR, 2017, pp. 1273--1282.

\bibitem{tan2022towards}
A.~Z. Tan, H.~Yu, L.~Cui, and Q.~Yang, ``Towards personalized federated learning,'' \emph{{IEEE} Trans. Neural Netw. Learn. Syst.}, 2022.

\bibitem{mansour2020three}
Y.~Mansour, M.~Mohri, J.~Ro, and A.~T. Suresh, ``Three approaches for personalization with applications to federated learning,'' \emph{arXiv preprint arXiv:2002.10619}, 2020.

\bibitem{smith2017federated}
V.~Smith, C.-K. Chiang, M.~Sanjabi, and A.~S. Talwalkar, ``Federated multi-task learning,'' \emph{Proc. Adv. Neural Inf. Process. Syst.}, vol.~30, 2017.

\bibitem{finn2017model}
C.~Finn, P.~Abbeel, and S.~Levine, ``Model-agnostic meta-learning for fast adaptation of deep networks,'' in \emph{Proc. Int. Conf. Mach. Learn.}\hskip 1em plus 0.5em minus 0.4em\relax PMLR, 2017, pp. 1126--1135.

\bibitem{danaher2014joint}
P.~Danaher, P.~Wang, and D.~M. Witten, ``The joint graphical lasso for inverse covariance estimation across multiple classes,'' \emph{J. R. Stat. Soc. B.}, vol.~76, no.~2, pp. 373--397, 2014.

\bibitem{bickel2008regularized}
P.~J. Bickel and E.~Levina, ``Regularized estimation of large covariance matrices,'' \emph{Ann. Statist}, vol.~36, no.~1, p. 199–227, 2008.

\bibitem{yuan2021joint}
Y.~Yuan, D.~W. Soh, X.~Yang, K.~Guo, and T.~Q. Quek, ``Joint network topology inference via structured fusion regularization,'' \emph{arXiv preprint arXiv:2103.03471}, 2021.

\bibitem{zhang2022time}
X.~Zhang and Q.~Wang, ``Time-varying graph learning under structured temporal priors,'' in \emph{Proc. Eur. Signal Process. Conf.}\hskip 1em plus 0.5em minus 0.4em\relax IEEE, 2022, pp. 2141--2145.

\bibitem{yamada2019time}
K.~Yamada, Y.~Tanaka, and A.~Ortega, ``Time-varying graph learning based on sparseness of temporal variation,'' in \emph{Proc. IEEE Int. Conf. Acoust., Speech, Signal Process.}\hskip 1em plus 0.5em minus 0.4em\relax IEEE, 2019, pp. 5411--5415.

\bibitem{zhang2024}
X.~Zhang and Q.~Wang, ``A graph-assisted framework for multiple graph learning,'' \emph{{IEEE} Trans. Signal. Inf. Process. Netw.}, pp. 1--16, 2024.

\bibitem{t2020personalized}
C.~T~Dinh, N.~Tran, and J.~Nguyen, ``Personalized federated learning with moreau envelopes,'' \emph{Proc. Adv. Neural Inf. Process. Syst.}, vol.~33, pp. 21\,394--21\,405, 2020.

\bibitem{bellet2018personalized}
A.~Bellet, R.~Guerraoui, M.~Taziki, and M.~Tommasi, ``Personalized and private peer-to-peer machine learning,'' in \emph{Proc. Int. Conf. Artif. Intell. Stat., AISTATS}.\hskip 1em plus 0.5em minus 0.4em\relax PMLR, 2018, pp. 473--481.

\bibitem{marfoq2021federated}
O.~Marfoq, G.~Neglia, A.~Bellet, L.~Kameni, and R.~Vidal, ``Federated multi-task learning under a mixture of distributions,'' \emph{Proc. Adv. Neural Inf. Process. Syst.}, vol.~34, pp. 15\,434--15\,447, 2021.

\bibitem{chen2022personalized}
F.~Chen, G.~Long, Z.~Wu, T.~Zhou, and J.~Jiang, ``Personalized federated learning with graph,'' \emph{arXiv preprint arXiv:2203.00829}, 2022.

\bibitem{stankovic2019introduction}
L.~Stankovi{\'c}, M.~Dakovi{\'c}, and E.~Sejdi{\'c}, ``Introduction to graph signal processing,'' in \emph{Vertex-Frequency Analysis of Graph Signals}.\hskip 1em plus 0.5em minus 0.4em\relax Springer, 2019, pp. 3--108.

\bibitem{hu2020multi}
Z.~Hu, F.~Nie, W.~Chang, S.~Hao, R.~Wang, and X.~Li, ``Multi-view spectral clustering via sparse graph learning,'' \emph{Neurocomputing}, vol. 384, pp. 1--10, 2020.

\bibitem{nie2017self}
F.~Nie, J.~Li, X.~Li \emph{et~al.}, ``Self-weighted multiview clustering with multiple graphs.'' in \emph{Int. Joint Conf. Artif. Intell.}, 2017, pp. 2564--2570.

\bibitem{hara2013learning}
S.~Hara and T.~Washio, ``Learning a common substructure of multiple graphical gaussian models,'' \emph{Neur. Netw.}, vol.~38, pp. 23--38, 2013.

\bibitem{lee2015joint}
W.~Lee and Y.~Liu, ``Joint estimation of multiple precision matrices with common structures,'' \emph{J. Mach. Learn. Res.}, vol.~16, no.~1, pp. 1035--1062, 2015.

\bibitem{karaaslanli2021multiview}
A.~Karaaslanli, S.~Saha, S.~Aviyente, and T.~Maiti, ``Multiview graph learning for single-cell rna sequencing data,'' \emph{bioRxiv}, 2021.

\bibitem{karaaslanli2024multiview}
A.~Karaaslanli and S.~Aviyente, ``Multiview graph learning with consensus graph,'' \emph{arXiv preprint arXiv:2401.13769}, 2024.

\bibitem{nesterov2013introductory}
Y.~Nesterov, \emph{Introductory lectures on convex optimization: A basic course}.\hskip 1em plus 0.5em minus 0.4em\relax Springer Science \& Business Media, 2013, vol.~87.

\bibitem{shokri2017membership}
R.~Shokri, M.~Stronati, C.~Song, and V.~Shmatikov, ``Membership inference attacks against machine learning models,'' in \emph{Proc. IEEE Symp. Secur. Privacy (SP)}.\hskip 1em plus 0.5em minus 0.4em\relax IEEE, 2017, pp. 3--18.

\bibitem{al2016reconstruction}
M.~Al-Rubaie and J.~M. Chang, ``Reconstruction attacks against mobile-based continuous authentication systems in the cloud,'' \emph{{IEEE} Trans. Inf. Forensics Security}, vol.~11, no.~12, pp. 2648--2663, 2016.

\bibitem{dwork2006calibrating}
C.~Dwork, F.~McSherry, K.~Nissim, and A.~Smith, ``Calibrating noise to sensitivity in private data analysis,'' in \emph{Proc. Theory of Cryptography Conf.}\hskip 1em plus 0.5em minus 0.4em\relax Springer, 2006, pp. 265--284.

\bibitem{li2020federated1}
T.~Li, A.~K. Sahu, M.~Zaheer, M.~Sanjabi, A.~Talwalkar, and V.~Smith, ``Federated optimization in heterogeneous networks,'' \emph{Proceedings of Mach. Learn. and Sys.}, vol.~2, pp. 429--450, 2020.

\bibitem{liu2022privacy}
K.~Liu, S.~Hu, S.~Z. Wu, and V.~Smith, ``On privacy and personalization in cross-silo federated learning,'' \emph{Proc. Adv. Neural Inf. Process. Syst.}, vol.~35, pp. 5925--5940, 2022.

\bibitem{li2021ditto}
T.~Li, S.~Hu, A.~Beirami, and V.~Smith, ``Ditto: Fair and robust federated learning through personalization,'' in \emph{Proc. Int. Conf. Mach. Learn.}\hskip 1em plus 0.5em minus 0.4em\relax PMLR, 2021, pp. 6357--6368.

\bibitem{fortunato2010community}
S.~Fortunato, ``Community detection in graphs,'' \emph{Phys. Rep.}, vol. 486, no. 3-5, pp. 75--174, 2010.

\bibitem{han2011data}
J.~Han, J.~Pei, and M.~Kamber, \emph{Data mining: concepts and techniques}.\hskip 1em plus 0.5em minus 0.4em\relax Elsevier, 2011.

\bibitem{kana2011disrupted}
R.~K. Kana, L.~E. Libero, and M.~S. Moore, ``Disrupted cortical connectivity theory as an explanatory model for autism spectrum disorders,'' \emph{Phys. Life Rev}, vol.~8, no.~4, pp. 410--437, 2011.

\bibitem{wallace1959bounds}
D.~L. Wallace, ``Bounds on normal approximations to student's and the chi-square distributions,'' \emph{Ann. Inst. Stat. Math.}, pp. 1121--1130, 1959.

\bibitem{saboksayr2021online}
S.~S. Saboksayr, G.~Mateos, and M.~Cetin, ``Online discriminative graph learning from multi-class smooth signals,'' \emph{Signal Process.}, vol. 186, p. 108101, 2021.

\bibitem{nie2010efficient}
F.~Nie, H.~Huang, X.~Cai, and C.~Ding, ``Efficient and robust feature selection via joint $\ell_{2,1}$-norms minimization,'' \emph{Proc. Adv. Neural Inf. Process. Syst.}, vol.~23, 2010.

\end{thebibliography}

\begin{comment}
\clearpage
\centering{\Large{\textbf{Supplementary Materials}}}
\appendices
\input{Supplementary/supplementary}
\end{comment}

\end{document}